\documentclass[sts]{imsart}

\RequirePackage{amsthm,amsmath,amsfonts,amssymb}
\RequirePackage[authoryear]{natbib}
\RequirePackage{graphicx}

\usepackage{algpseudocode}
\usepackage{algorithm}

\newcommand{\specialcell}[2][c]{%
\begin{tabular}[#1]{@{}c@{}}#2\end{tabular}}

\algblockdefx[Train]{Train}{EndTrain}[1][]{\textbf{Training} #1}{\textbf{End Training}}
\algblockdefx[Operation]{Operation}{EndOperation}[1][]{\textbf{Operation} #1}{\textbf{End Operation}}

\usepackage[utf8]{inputenc}
\newcommand{\argmax}{\operatorname*{\arg\max}}

\newcommand{\dd}[1]{\text{d}#1}

\usepackage[colorlinks=true,allcolors=black,breaklinks=true]{hyperref}
\usepackage{multicol}
\usepackage{multirow}
\usepackage{tabularx}
\usepackage{booktabs}
\usepackage{algpseudocode}
\usepackage{hyperref}
\usepackage{subcaption}

\startlocaldefs

\endlocaldefs

\begin{document}

\begin{frontmatter}
\title{Protecting Classifiers From Attacks}
\runtitle{Protecting Classifiers From Attacks}

\begin{aug}
\author[A]{\fnms{V\'ictor} \snm{Gallego}}\footnote{Both authors contributed equally.}$^{a}$,
\author[B]{\fnms{Roi} \snm{Naveiro}}\footnotemark[\value{footnote}]$^{b}$,
\author[C]{\fnms{Alberto} \snm{Redondo}}$^c$,
\author[D]{\fnms{David} \snm{Ríos Insua}}$^c$
\and
\author[E]{\fnms{Fabrizio} \snm{Ruggeri}}$^d$
\address[A]{$~^a$Komorebi AI Technologies.}
\address[B]{$~^b$CUNEF Universidad.}
\address[C,D]{$^c$Institute of Mathematical Sciences (ICMAT-CSIC).}
\address[E]{$~^d$CNR-IMATI.}
\address[F]{$~^*$Both authors contributed equally.}
\end{aug}

\begin{abstract}
In multiple domains such as malware detection, automated driving systems, or fraud detection, classification
algorithms are susceptible to being attacked by malicious  agents willing to perturb the value of instance covariates to pursue 
certain goals. Such problems pertain to the  field of adversarial machine learning and have been mainly dealt with,  perhaps implicitly,  through game-theoretic ideas with strong underlying common knowledge assumptions. These are not realistic in
numerous application domains in relation to security
and business competition. We present  an alternative Bayesian decision theoretic  
framework that accounts for the uncertainty about the attacker’s behavior using adversarial risk analysis
concepts. In doing so, we also present core ideas in adversarial machine learning to a statistical audience. A key ingredient
in our framework is the ability to sample from the distribution of originating instances given the, possibly attacked, observed ones. We propose an initial procedure based on approximate Bayesian computation usable during operations; within it, we simulate the attacker’s problem taking into account our uncertainty about his elements. Large-scale problems require an alternative scalable approach implementable during the training stage. Globally, we are able to robustify statistical classification algorithms against malicious attacks.
\end{abstract}


\end{frontmatter}
\section{Introduction}
\label{sec:introduction}

Over this and the last decade, an increasing number of processes is being 
automated through classification algorithms \citep{bishop2006pattern}.
It is thus essential that these are robust in order to trust key operations based on their output. As a fundamental hypothesis, statistical classification relies on the use of independent and identically distributed (iid) data for both the training and operation phases. State-of-the-art classifiers perform extraordinarily well on such data, but they have proved vulnerable to various types of attacks targeted at fooling the underlying algorithms \citep{comiter,trend}.
Security aspects of classification, which form part of the emerging field of {\em adversarial machine learning} (AML) \citep{vorobeichikantar,joseph}, question the iid hypothesis due to the presence of adversaries ready to modify the data to obtain a benefit, making training and operation distributions different. 
 The societal relevance of the problem is well-reflected in the recently proposed EU Artificial 
Intelligence (AI) Act \citep{EU}, NIST AI Risk Management Framework \citep{nist} 
and NIST-MITRE AML terminology \citep{mitre}.

Work in AML has traditionally focused around 
three topics: a) analyzing  {\em attacks} against machine learning (ML) algorithms to
 uncover their vulnerabilities; 
b) delivering {\em defenses} against such attacks; and, c) consequently,
developing {\em frameworks} encompassing both attacks and defenses.
While an important methodological pillar in the area, see e.g.\
\cite{joseph}, has been classical robust statistics 
\citep{hampel}, the predominant paradigm used to frame the confrontation between classification systems and adversaries has been,
 sometimes implicitly, game theory (see reviews in \cite{BIGGIO2018317} and \cite{doi:10.1002/widm.1259}). 
As examples, a) the most popular attacks, including the fast gradient sign method (FGSM) \citep{goodfellow2014explaining}, may be viewed in game-theoretic terms as best responses 
(i.e. maximizing the adversary's utility) to the 
classification algorithms; similarly, b),
one of the most promising defense techniques, \emph{adversarial training} (AT) 
 \citep{madry2018towards}, 
may be framed  through 
best response defenses against a worst-case attack (i.e. those maximizing  the defender's utility under the worst-case attack); and, c), finally, 
the pioneering framework in adversarial classification (AC)   \citep{adversarialClassification2004} was introduced as a game between a classifier
and an adversary. 
This perspective typically entails strong common knowledge (CK) assumptions \citep{gameTheoryACriticalIntroduction2004} which, from a fundamental point of view, are not sustainable in important application domains like security, defense, law enforcement or 
competitive business, as adversaries attempt to hide and conceal information.

\textcolor{black}{Recent work \citep{naveiro2018adversarial} presented ACRA, a novel approach to AC based on Adversarial Risk Analysis (ARA) \citep{adversarialRiskAnalysis2009}.} 
ARA makes operational a Bayesian approach to games (as in \cite{kadane1982subjective} and \cite{raiffa1982art}) facilitating procedures to predict adversarial decisions \textcolor{black}{  used to enhance the 
robustness of classifiers.} However, ACRA may be used only with generative classifiers \citep{10.5555/3086952}, like \textcolor{black}{ the utility sensitive Naive Bayes classifier \citep{chai2004test}
or certain deep variational autoencoder classifiers \citep{kingma2014semi}. Moreover, only binary classification problems were supported in ACRA.}

\textcolor{black}{ Thus, besides reviewing core AML ideas  
relevant in AC, we  present a general \textcolor{black}{ Bayesian 
decision theoretic} framework }  that 
may be used with both discriminative and generative classifiers,
deals with multiple class problems and provide efficient
computational schemes in large scale \textcolor{black}{settings}. 
In this, we not only solve a current complex problem with a sophisticated solution but also
aim to bring the attention of the statistical community to a very relevant and largely unexplored issue: performing statistical inference in presence of adversaries. This 
problem area is not only a matter of academic interest but also a serious concern for individual and societal security, \textcolor{black}
{ as expressed in the recent {\em Executive Order on the Safe, Secure, and Trustworthy Development and Use of Artificial Intelligence} \citep{whitehouse} }.
The dramatic increase in data availability has made classification an even more urgent need, as well as that of protecting from misclassification, either caused by intentional agents or by nature. ML  techniques are getting increasingly relevant and, thus, AML is gaining track   within the computer science community (not so much among statisticians). Thus, our purpose is to \textcolor{black}{ stir leveraging statistical techniques to tackle core problems in AML}.

With this in mind, \textcolor{black}{ a broad overview of the general AC problem 
structures this paper according to the three core topics mentioned above}.
First (topic a), Section 2 provides a setup of the general AC problem, 
showcases two examples of how the performance of classifiers degrades in 
presence of subtle attacks, and overviews key attacks.
Then (topic b), Section 3 overviews state-of-the-art defenses and 
suggests  a general Bayesian solution to robustify classifiers against adversarial data manipulations.  This approach changes the way classification decisions are made {\em during operations} 
and we illustrate its performance in a sentiment analysis problem; 
computational issues are discussed and an efficient alternative for large-scale problems is proposed, affecting the {\em training stage}, and modifying the way inferences are performed to take into account the eventual presence of adversaries during operations. It is illustrated through a case in computer vision based on a deep neural network classifier. \textcolor{black}{ Finally (topic c), a  computational pipeline 
encompassing the whole framework is presented in Section 4.}
\textcolor{black}{ Code to reproduce all the experiments in the paper and 
illustrate the proposed pipeline is available at \url{https://github.com/datalab-icmat/aml_bayes_classification}}.

\section{ Attacks over classifiers}\label{sec:cla_att}

\subsection{Basic setup}
Consider a classifier $C$ (she) which may receive instances 
from $k$ different classes designated with a label $y \in \lbrace 1, \dots, k \rbrace$. Instances have covariates/features $x \in \mathbb{R}^d$. 
 \textcolor{black}{  Notationwise, for convenience, we distinguish between random variables and realizations using upper and lower cases, respectively; thus
 $X=x$ refers to the originating covariates.}
Uncertainty about the instances' class given its covariates is modeled through a distribution $ p(y| x)$, usually parameterized with certain parameters $\beta$. 
  Such distribution can come from a generative model, 
\textcolor{black}{ e.g.\ a naive Bayes one}, where distributions $p(x)$ and $p(x|y)$ are explicitly modeled and $ p(y| x)$ is obtained through  Bayes formula; or from a discriminative model, \textcolor{black}{e.g., a neural
network}, in which $ p(y| x)$ is modeled directly \citep{bishop2006pattern}. 
{\color{black}
The $\beta$ parameters are estimated using training data $\mathcal{D} = \left \lbrace (x_i, y_i)_{i=1}^N \right \rbrace$. In classical approaches, data $\mathcal{D}$ is used to find an estimate $\hat{\beta}$ that maximizes a certain likelihood (or minimizes a loss function $L(\beta, x, y)$)  and $p(y|\hat{\beta},x)$ is employed for classifying new instances. In Bayesian approaches, a prior $p (\beta)$ is used to compute the posterior $p (\beta | \mathcal{D} )$ and the predictive distribution $p(y | x, \mathcal{D})$ is used to classify new data. Herein, $p(y|x)$ refers to either the predictive distribution of $y$ when evaluating covariates $x$ if we are in a Bayesian setup or $p(y_i|\hat{\beta},x)$ if we are in the classical one.}
%
Other non-probabilistic classifiers like support vector machines (SVMs) adapt as well to the above notation using calibration
\textcolor{black}{
methods such as 
\cite{platt1999probabilistic} scaling scheme}.

Whatever the estimation method adopted, $C$ aims at 
classifying $x$ to pertain to the class 
defined through $\,\, \arg\max_{y_C} \sum_{y=1}^k u_C (y_C , y ) p (y | x )$, where $u_C (y_C, y)$ is the utility that she perceives when an instance with label $y$ is classified as of class $y_C$,
thus using the maximum expected utility principle \citep{french2000statistical}.
%
%
\textcolor{black}{ Quite often, in the classification domain a   
$0-1$ utility function is used, so that } the classifier gets utility 1 for a
\textcolor{black}{
correct classification, and 0 for an incorrect one}. In this case, the decision rule is $\,\, \argmax_{y_C} 
p ( y = y_C | x )$ and we thus aim to \textcolor{black}{find the class that maximizes the probability of a } correct classification. 

\textcolor{black}{In the scenario of interest, another agent, called adversary $A$ (he),
is involved. He applies an attack $a$ to the features $x$ leading to the transformed covariates $x'=a(x)$  actually received by $C$.} 
The adversary aims to fool the classifier by making her misclassify instances to attain some benefit as it happens, e.g., in spam detection,   
 where, by crafting his message, a spammer aims to fool a spam 
detection tool so as to make her classify a 
spam message as \textcolor{black}{legitimate} to increase his business opportunities.
\textcolor{black}{Upon observing $x'$, $C$ needs to determine the instance class.
An adversary unaware classifier might be making gross mistakes as she classifies based on the received, possibly modified, features $x'$, instead of the actual ones, which are not observed.
We provide two societally relevant examples that 
will drive our proposals in Sections 3.2 and 3.3, respectively.}
\\

{\color{black}
\subparagraph*{Example 1.} 
Consider a sentiment analysis problem. The goal is to assess whether a film review was positive or negative. We use a dataset containing 2400 IMDb reviews (1200 positive, 1200 negative) extracted from \cite{kotzias2015group}. As covariates,  we use 150 binary features indicating the presence or absence of the most common 150 words in the dataset 
 \textcolor{black}{ after removing stopwords. A label } indicates whether the review is positive ($y = 0$) or negative ($y = 1$). 

We study the performance degradation of \textcolor{black}{four standard statistical classification algorithms ({\em  naive Bayes, logistic regression (LR), neural network (NN)} and {\em  random forests}) 
under the actions of two types of attackers}. For the first one, referred to as \textit{attacker A}, the adversary 
aims to manipulate positive reviews in such a way that they are classified as negative, thus artificially decreasing the predicted quality of the film.
The goal of the second adversary, denoted \textit{attacker B}, is to manipulate negative reviews so as to make the classifier label them as positive ones,  introducing bad reviews without being noticed.
In both cases, we consider an attacker that modifies reviews by either adding or removing words. The perturbations have to be somehow restricted for the reviews to conceal 
their malicious intent. We do so by allowing at most two modifications per review.

Table \ref{tab:cleanVSattack} presents the accuracy
of the four classifiers  over clean and attacked test data; LR is applied with L2 regularisation (equivalent to performing maximum a posteriori (MAP) estimation in an LR model with a normal prior); the NN has two hidden layers.  
All models are trained under the same conditions: we randomly split the dataset into train and test subsets, respectively, with sizes $90\%$, $10\%$. 
Accuracy means and standard deviations are estimated via hold-out validation over 10 repetitions \citep{kim2009estimating}.
\begin{table*}[htb]

\caption{Accuracy comparison (with precision) 
 	of four classifiers on clean and attacked data.}
\centering
\begin{tabular}[t]{llll}
\toprule
\textbf{Classifier} & \textbf{Clean data} & \specialcell{\textbf{Attacked data}\\ Attacker A} & \specialcell{\textbf{Attacked data}\\ Attacker B}\\
\midrule
Logistic Regression & $ 0.728 \pm 0.005 $ & $ 0.322 \pm 0.011 $ & $ 0.418 \pm 0.010 $\\
Naive Bayes & $ 0.722 \pm 0.004 $ & $ 0.333 \pm 0.009 $ & $ 0.405 \pm 0.009 $\\
Neural Network & $ 0.691 \pm 0.019 $ & $ 0.338 \pm 0.021 $ & $ 0.417 \pm 0.015 $\\
Random Forest & $ 0.720 \pm 0.005 $ & $ 0.327 \pm 0.011 $ & $ 0.397 \pm 0.013 $\\
\bottomrule
\end{tabular}
\label{tab:cleanVSattack}%
\end{table*}
  Clearly, all four classifiers experience a 
considerable performance degradation that highlights their lack of robustness against adversarial attacks. \hfill $\triangle$ 
\\

}

\subparagraph*{Example 2.}
\textcolor{black}{ The second experiment is in the computer vision domain.
It refers to a handwritten digit 
recognition problem based on a deep neural network classifier 
with $k = 10$ classes (1 per digit)}.
The best-known attacks to classification algorithms in this 
domain, \textcolor{black}{ like the above mentioned FGSM,}
purposefully modify images so that alterations become imperceptible to the human eye, yet drive a model to misclassify the perturbed ones 
\textcolor{black}{ greatly degrading performance}. These perturbed images are denominated 
{\em adversarial examples} \citep{Szegedy14intriguingproperties}. 

\begin{figure*}[hbt]
\centering
\begin{subfigure}{.3\textwidth}
  \centering
  \includegraphics[width=.7\linewidth]{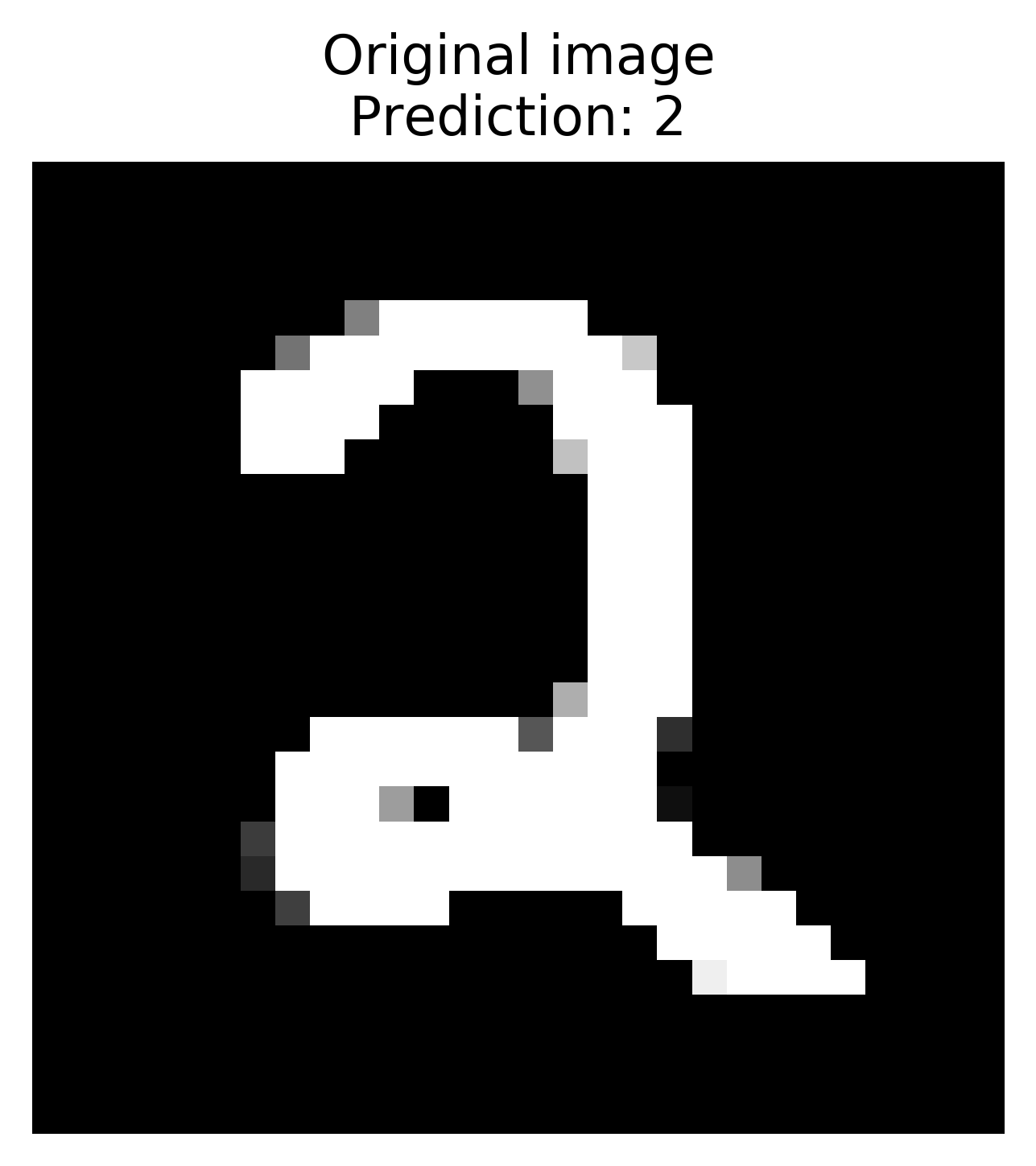}
  \caption{Original image}
  \label{fig:2}
\end{subfigure}%
\begin{subfigure}{.3\textwidth}
  \centering
  \includegraphics[width=.7\linewidth]{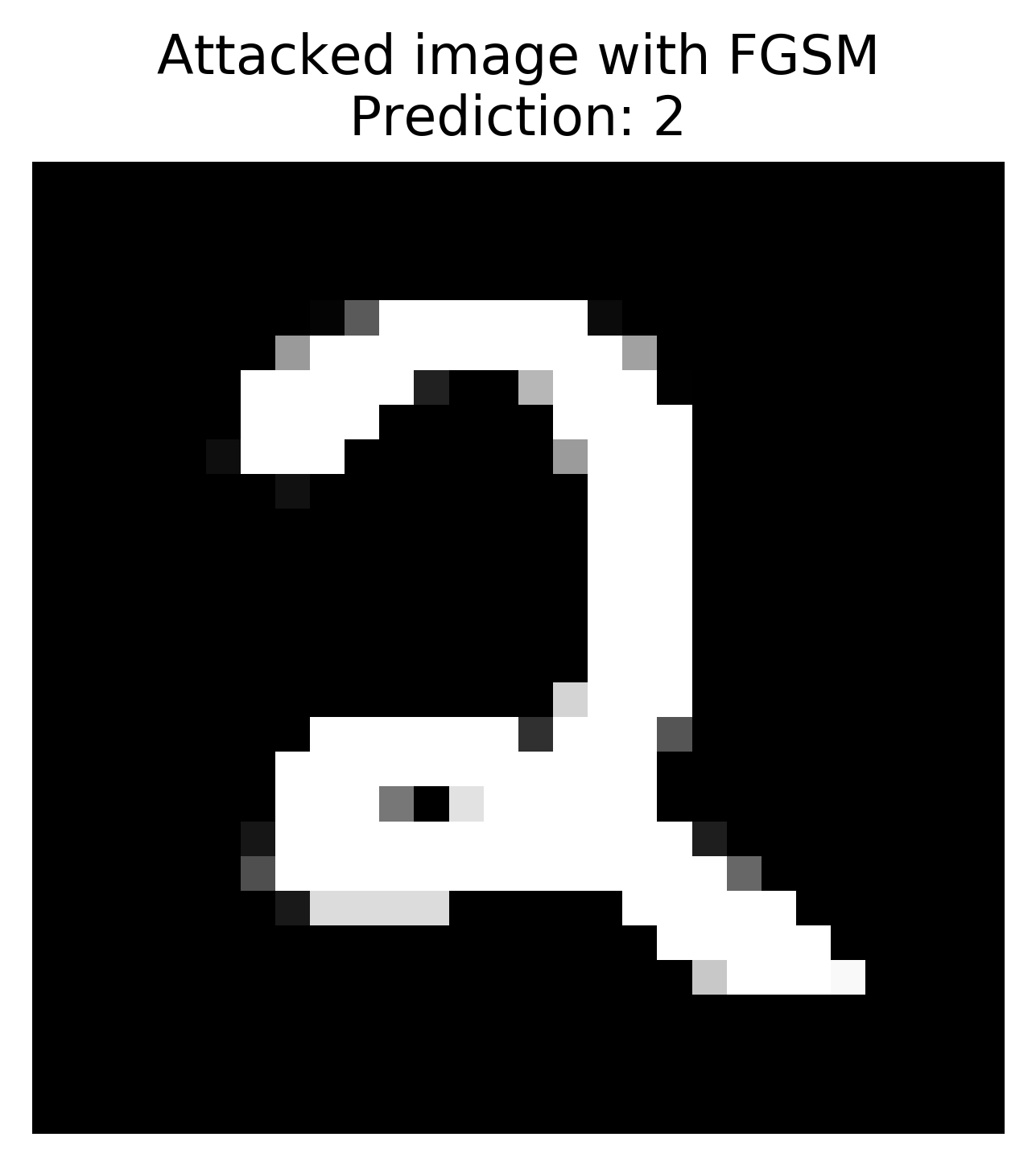}
  \caption{FGSM-attacked image}
  \label{fig:2b}
\end{subfigure}
\begin{subfigure}{.3\textwidth}
  \centering
  \includegraphics[width=.7\linewidth]{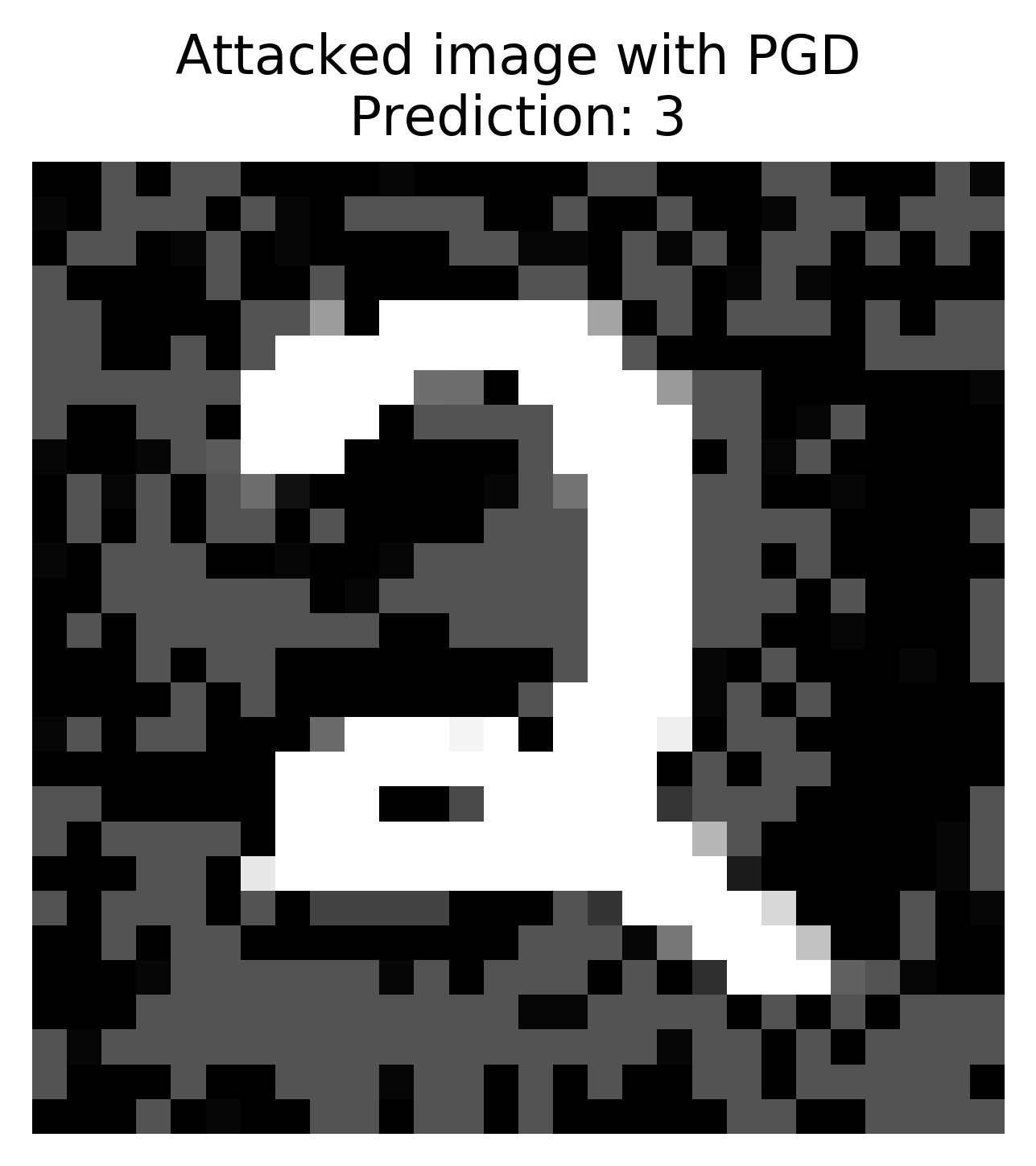}
  \caption{PGD-attacked image}
  \label{fig:2c}
\end{subfigure}
\caption{An original MNIST image identified as 2. \textcolor{black}{The same image, perturbed with FGSM. Some pixels have changed, but it is still correctly classified. The same image, perturbed with PGD, is now incorrectly classified as a 3}.}
\label{fig:27}
\end{figure*}

  For instance, with a relatively simple deep convolutional neural network (CNN) model \citep{krizhevsky2012imagenet}, we \textcolor{black}{achieve 99\% out of the sample accuracy when predicting the handwritten digits in the MNIST data set \citep{MNIST} with $28\times 28$ pixels (thus, $d=784$ features are used).} 
However, if we perturb such a set with the FGSM attack, the accuracy gets drastically reduced to 62\%. \textcolor{black}{Moreover, if we use a more powerful attack such as the Projected Gradient Descent (PGD) method, accuracy rapidly decays to practically 0\%.
Figure \ref{fig:27} provides an example of (a) an original image, 
  (b) a perturbed one with FGSM, and (c) a perturbed one with PGD. To our eyes, the three images look like a 2. However, with high probability our CNN classifier correctly identifies a 2 in the first two 
 cases (Fig. \ref{fig:2} and \ref{fig:2b}), yet classifies as a 3 that under 
 a PGD attack (Fig. \ref{fig:2c}), .}


\noindent 
\textcolor{black}{ A direct analog of this example in the 
automated driving system setup relates to the confounding
of a yield sign with a stop sign \citep{ASMBI}, with 
potentially catastrophic consequences.} \hfill $\triangle$

\subsection{Overview on attacks}\label{attacks}

{\color{black}
The FGSM attack in Example 2 is defined through
\begin{equation*}
    x' = x + \epsilon \cdot \text{sign} (\nabla_x L(\beta,x,y) )
\end{equation*}
where $x'$ designates the attacked covariates; $x$ are the original ones;
 $\nabla_x L(\beta , x,y)$ is the gradient of the loss function with respect to $x$; and  $\epsilon$ is a small scalar reflecting attack intensity. 
Thus, the FGSM attack can be interpreted as a best response when the attacker can perturb each covariate by $\epsilon$.  Importantly,
 observe that this attack assumes that the adversary has precise knowledge of
the underlying model and parameters of the involved classifier,
what is designated a {\em white box} attack \citep{mitre}.

Another well-known white box attack is the PGD method \citep{madry2018towards}
which iterates through the expression \textcolor{black}{until the loss function plateus}
\begin{equation}\label{monday1}
x_{t+1} =  \text{Clip}_{x, \epsilon} \left \lbrace x_t + \alpha \cdot \text{sign} (\nabla_x L(\beta,x_t,y_t) )
\right \rbrace
\end{equation}
where $\alpha$ is the gradient step size and the Clip function forces the distance between the new instance covariates and their original values to be less than $\epsilon$. Finally,  \cite{carlini2017towards} 
white box attack has gained popularity recently. It looks for perturbed instances $x' = x + \epsilon$, where the perturbation is obtained through
\begin{eqnarray*}
\min_{\epsilon} \Vert \epsilon \Vert_p + c \cdot f_y(x+\epsilon),
\end{eqnarray*}
with $\Vert \cdot \Vert_p$ being the $L_p$ norm and $f_y(x + \epsilon)$ is a function dependent on the loss $L(\beta, x, y)$ such that $f_y(x + \epsilon) \leq 0$ if and only if the output class for $x + \epsilon$ is the target class $y$. The positive constant $c$ trades-off between minimizing the norm of the perturbation and maximizing its adversarial effect.

Notice that minimizing the most common loss functions in ML is equivalent to maximizing a certain posterior distribution since such 
loss functions can be typically written as
%
%
%
\begin{equation} \label{aux}
     \sum_{i=1}^N L(\beta, x_i, y_i) = - \sum_{i=1}^N \log p(y_i \vert x_i, \beta ) - \log p(\beta),
\end{equation}
where $L(\beta, x_i, y_i)$ is the loss function evaluated at training instance $(x_i, y_i)$.
For instance, in a logistic regression setting with normal priors over the coefficients, finding their MAP estimate is equivalent to minimizing the cross-entropy loss with $L_2$ regularization. Thus, all the reviewed attacks can be seen as approximations of the optimization problem
$$
x'= \arg\min_{z  \in B(x)} \log p(y| z,\beta),
$$ 
where $B(x)$ is some neighborhood of $x$ 
over which the attacker has an influence.
The exact solution to this problem is intractable in high-dimensional data
and thus we resort to the above type of approximations based on gradient information.
However, they all assume that the attacker has full knowledge of the target model,
which is unrealistic in most security scenarios, except for the case of insiders
 \citep{joshi}.}

{\color{black}
The type of attacks in 
Example 1
demands much less information from the defender model, but still some as it requires to know, e.g., 
the features used by the classifier. Thus, it corresponds to a {\em gray box }
attack \citep{mitre} in which the classification system is partially known by the attacker.
However, the entailed assumptions might be still debatable in most security domains. 

Finally, there are {\em black box } attacks \citep{mitre}
which make minimal assumptions about the classifier.
One example is the good word insertions attack against spam 
detectors in \cite{naveiro2018adversarial}
Another instance are the one-pixel attacks in \cite{su}.
The entailed assumptions are 
reasonable, yet their applicability is limited
in numerous setups. 

\textcolor{black}{The above colored attack terminology is based on the knowledge that the attacker might have about the classifier.  There are other taxonomies;  for example, in \cite{RiosInsua2018} adversaries 
are distinguished according to their capabilities or the type
of problem they solve as {\em data fiddler}, {\em structural} and 
{\em parallel} attackers.
\cite{AdversarialMachineLearning2011,Barreno2006} and \cite{mitre} 
 provide further attack taxonomies } based on attackers' goals, 
knowledge and 
capabilities and attack
timing.

In this paper, we focus on exploratory attacks (defined to have influence just over operational data, leaving training data untainted). Moreover, we shall consider only integrity violations (the adversary just modifies malicious instances 
trying to make them be classified as legitimate ones). This is the most common case in 
numerous application domains. For instance, in fraud detection, fraudsters might modify the way their operations are done, trying to avoid them being classified as fraudulent by a classifier trained with clean data.

\textcolor{black}{ 
As a major driver of this paper, we shall emphasize 
modeling inherent uncertainty about attacks: in realistic settings, 
 the classifier would not have precise information about how the attacker modifies a given instance, as, in general, his preference and probability assessments are 
 not fully available. It seems therefore crucial to account explicitly for such uncertainty. }
}

\section{Protecting classifiers from attacks}
\label{sec:defenses}

\subsection{Incorporating defenses}

\textcolor{black} {Examples 1 and 2 showcased that an adversary unaware classifier may be purposefully} fooled into issuing wrong classifications potentially incurring in severe expected utility underachievement.
Assume for a moment that the classifier knows the attack $a$ that she has suffered
and that such attack is invertible, in the sense that she may recover the original $x$, designated $a^{-1}(x')$ when convenient. Then, rather than classifying based on $\,\,\, \arg\max_{y_C} \sum_{y=1}^k  u (y_C , y ) p (y | x '), \,\,$ as an adversary unaware classifier would do, 
she should classify using 
\begin{equation}\label{monay2}
\arg\max_{y_C} \sum_{y=1}^k  u (y_C , y ) p (y | x=a^{-1} (x')).
\end{equation}
However, the classifier does not know the attack $a$, nor, more generally, the originating instance. The way around this issue entails constructing models of the attacks likely to be undertaken by the adversaries.

\textcolor{black}{
As the Introduction discussed, most of the previous work along these lines rely on game theoretic arguments and classical robust statistical concepts.} 
Standard adversarial robustness definitions, e.g. \citep{katz2017towards}, aim at the condition
\begin{equation}\label{monday3}
\arg\max_y p(y|x,\beta) \approx \arg\max_y p(y|x',\beta), 
\end{equation}  
entailing that for any input $x$ and adversarial perturbation $x' \sim p(\cdot |x,y)$, the predicted class should not practically change under an adversarial attack. \textcolor{black}{Optimization problems 
(\ref{monday3}) can be stated by maximizing the log-likelihoods of the model under clear and attacked data,
that is $\max \log p(y|x, \beta)$ and $\max \log p(y|x', \beta)$}.
In general, methods to increase model robustness against adversarial data manipulations fall into two main categories: those that affect operations, modifying the rule used to classify new instances, and those that alter the inferences made in the training phase to take into account the eventual presence of adversaries during operations.

\textcolor{black}{ 
The pioneering work of \cite{adversarialClassification2004} is an example of the first class of methods. This work assumes that the classifier has full knowledge of the adversary's problem. Next, the attacked covariates produced when the attacker receives a given instance can be computed exactly. Finally, during operations, provided that we observe $x'$, the possible originating instances can be computed as those that, under the attack, lead to such observation. 
}

\textcolor{black}{
The aim of the second class of robustifying approaches is to train using artificial data that somehow mimic actual, potentially attacked, operational data, through several heuristic approaches. Most of these methods model how the attacker would modify the instances in the training set. AT  \citep{madry2018towards}  is a mainstream  proposal; instead of training a classifier to minimize the empirical risk based on training data, the authors minimize empirical risk under a worst-case attacker who chooses, for each instance, the worst modification within a constrained region. Having trained the classifier in this manner, $p (y | x')$ could be directly evaluated at the operational stage as this probability has been inferred taking into account the presence of an attacker.
\textcolor{black}{
AT can be viewed as a zero-sum game where complete knowledge about the adversary is assumed, since the attacker has total knowledge of the model's loss function and parameters, overlooking existing uncertainty. Although AT strives for worst-case guarantees, it may be less effective if the real adversary behaves differently than the assumed scenario. Recall that a main theme within this paper is that, by introducing the uncertainties faced by the adversary and the defender, we can expand upon AT to arrive at more robust and principled defenses.
}
}

Similarly, Adversarial Logit Pairing (ALP) defenses \citep{kannan2018adversarial} try to impose the stronger condition that the logits for clean and attacked instances are close,
\begin{equation}\label{tumorout}
p(y|x,\beta) \approx p (y|x',\beta),
\end{equation}
with improved robustness. This is done by including in the loss function minimized during training an extra term  proportional to the absolute difference of the logits. \textcolor{black}{Then problem to be solved is then $\min \lbrace - \log p(y|x, \beta) + \log p(y|x', \beta) + | f_\beta(x) - f_\beta(x') | \rbrace$, with $f_\beta(x)$ being the logits of the model under input $x$}.
As with AT, these defenses assume models for how the attacker would modify training instances that do not take into account the existing uncertainty, as for each training instance $x$, $x'$ is assumed to be available.



\subsection{Robustifying classifiers during operation}\label{sec:adv_aware_class}

This section analyses in detail the robustification of classifiers by modifying their operational stage. When observing instance $x'$, we model our uncertainty about the latent originating instance $x$ through a distribution $p(x \vert x')$ and perform inference about $x$ proposing a formal way to sample  from such distribution, thus 
 accounting for the lack of knowledge about the attack.
 \textcolor{black}{ However,  sampling from this  distribution  
is 
harder, 
especially in large scale settings,  
  and Section \ref{sec:diff}} proposes an approach that bypasses this harder step, modifying instead 
the training stage. 

\textcolor{black}{ Suppose for the moment that } we are actually able to model our uncertainty about the originating covariates $x$ given the observed $x'$ through a distribution $p(x|x')$ with support ${\cal X}_{x'}$, the set of reasonable instances $x$ leading to $x'$ if attacked. Based on 
(\ref{monay2}), the expected utility that
the classifier would get for a classification decision $y_C$ would be 
\begin{eqnarray*}
\psi (y_C) &=& \int _{{\cal X}_{x'}}
 \left( \sum_{y=1}^k u (y_C , y ) p (y | x=a^{-1} (x'))\right)
 p (x | x' ) dx  \\
 &=&  \sum_{y=1}^k   u (y_C , y ) \left[\int_{{\cal X}_{x'}} p (y | x=a^{-1} (x'))  p (x | x' ) dx \right], 
\end{eqnarray*}
 having to solve 
\begin{equation}\label{superpedo}
\arg\max_{y_C}\psi (y_C).
\end{equation} 
Typically, we approximate expected utilities by Monte Carlo (MC) using a sample $\{ x_n \}_{n=1}^N$ from $p(x | x' )$ so that 
$$ 
\widehat {\psi } (y_C)
=  \frac{1}{N} \sum_{y=1}^k  u (y_C , y ) \left[\sum _{n=1}^N p (y | x_n ) 
\right].
$$
As a consequence, Algorithm \ref{alg:gen_ara} summarizes \textcolor{black}{ a general procedure 
for adversarial classification } that we later specify.   

\begin{algorithm}[h] 
\caption{ARA procedure for Adversarial Classification during operations}  
\label{alg:gen_ara}
\begin{algorithmic}
\State {\bf Input:} $N$, training data $\mathcal{D}$.
\State {\bf Output:} A classification decision $y_C^*(x')$.
\Train
\State Based on ${\cal D}$ 
estimate a model for $p(y|x)$. 
\EndTrain
\Operation
\State Read instance $x'$
\State Draw sample $\{ x_n \}_{n=1}^N$ from $p(x|x')$.\\
\State Find  
{\small 
$y_C^* (x')= \argmax_{y_C} \frac{1}{N}\sum_{y=1}^k   \left(u (y_C , y ) \left[ \sum _{n=1}^N p (y | x_n ) \right] \right)
$ } \\
\EndOperation
\State \textbf{Return} $y_C^* (x')$
\end{algorithmic}
\end{algorithm}

 To implement this approach, inference about the latent originating instance $x$ given the observed $x'$ must be undertaken. This entails estimating $p(x|x')$ or, at least, being able to sample from it. To do so, one must define an \textit{attack model} $p(x'|x)$, that is, a model of our beliefs about how the attacker modified instance $x$, and sample from it. From a modeling perspective, this is involved as it requires strategic thinking about the adversary. Later, we suggest a formal Bayesian decision-theoretic argument to produce such a sample. 
For the moment, assume that such a procedure is available. \textcolor{black}{   Note that, if we could evaluate $p(x'|x)$ and $p(x)$ analytically, then sampling from $p(x|x')$ could be done using standard MCMC methods \citep{french2000statistical}. However, in general, estimating an \textit{attacking model} is much harder than simulating from it.} 

\subsubsection{\textbf{A specification: AB-ACRA} }

\textcolor{black}{ We propose AB-ACRA, an approach to sample
from $p(x | x')$  making use of samples from $p(x'|x)$ by } leveraging the information available about the attacker using concepts from approximate Bayesian computation  (ABC) \citep{csillery2010approximate} and ARA 
 \citep{Banks}. As basic ingredients, the approach requires sampling from $x \sim p(x)$ and $x' \sim p(x' | x)$. 

\textcolor{black}{ Assume initially that $x$ and, thus, $x'$ are discrete. }
In this case, we could easily generate samples $p(X | X' = x')$ using MCMC including a rejection step. This would entail proposing a candidate $\tilde{x}$ according to certain transition distribution $q(x \rightarrow \tilde{x})$, sampling $\tilde{x}' \sim p(X' | X= \tilde{x})$ and, if the generated $\tilde{x}'$ is equal to the instance $x'$ actually observed by the classifier, accept $\tilde{x}$ with probability $ \alpha = \min \left \lbrace 1, \frac{p(\tilde{x}) q(\tilde{x} \rightarrow x_i)}{p(x_i) q(x_i \rightarrow \tilde{x})}\right \rbrace$. Using standard reversibility arguments in Metropolis-Hastings algorithms, it is straightforward to prove that samples generated iterating through these steps are approximately 
distributed according to $p(X | X' = x')$.
%

%
Note a few things. First, this approach requires us to evaluate $p(X)$. If this is not possible, but we can generate samples from such distribution, we could choose $p(X)$ to be the proposal generating density $q$. It can be easily seen that, in this case, the acceptance probability is just
$\mathbb{I}[\tilde{x}' = x']$, thus avoiding the evaluation of $p(X)$. Indeed, one iteration of the previous scheme will produce a sample from the desired distribution, being an instance of a rejection sampler \citep{casella}. Note that
we would be generating $\tilde{x} \sim p(X)$, $\tilde{x}' \sim p(X' | X = \tilde{x})$, and accepting $\tilde{x}$ only if $\tilde{x}'$ coincides with the actually observed instance $x'$. It is straightforward to prove that $\tilde{x} \sim p(X | X' = x')$: think of this procedure as generating instances $x$ and indicators $I$, where $I=0 (1)$ if we reject (accept) the sample; then,
accepted instances are distributed according to
\textcolor{black}{ the required distribution since 
\begin{eqnarray*}
    p(X = \tilde{x} | I=1) &\propto& p(I=1 | X= \tilde{x}) p(X=\tilde{x}) \\
    &\propto& p(X' = x' \vert X=\tilde{x})p(X=\tilde{x}) \\ 
    &\propto&  p(X = \tilde{x} | X' = x').
\end{eqnarray*}
}
In general, however the convergence of this MCMC scheme will 
be slow, as just samples for which $\tilde{x}' = x'$ are  accepted. Speed is clearly affected by the choice of $q$: densities that produce instances $\tilde{x}$ such that $p(X' | X=\tilde{x})$ placing a lot of mass around the observed $x'$ will have better mixing. However, when $x'$ is high dimensional the acceptance rate would be very low, as $p(X'=x' | X)$ will be generally very small. Moreover, in the continuous case, it will be $p(X'=x' | X) = 0$, canceling the acceptance rate.

We thus leverage ABC techniques \textcolor{black}{ \citep{ABC}.
 For this, let us } relax the condition $\tilde{x}' = x'$ in the rejection step of the previous MCMC scheme, and allow samples
 \textcolor{black}{ which are sufficiently close in the sense that }
  $\phi(\tilde{x}', x') < \epsilon$ for a given distance function 
$\phi$ and tolerance $\epsilon$. Observe though that the probability of generating samples for which $\phi(\tilde{x}', x') < \epsilon $ decreases as the dimension of $x'$ increases. A common ABC solution replaces the acceptance criterion with the condition 
$\mathbb{I}\left[ \phi(s(\tilde{x}'), s(x') ) < \epsilon \right]$, where $s(x)$ is a set of summary statistics that capture the relevant information in $x$, the particular choice of $s$ being problem specific  \textcolor{black}{ (see
 examples for $\phi$ and $s$
in the case study in 3.3.2).} Obviously, if we use $\mathbb{I}\left[ s(\tilde{x}') = s(x') \right]$ the approach is exact, provided that $s(x)$ is a sufficient statistic for $X$ in $p(X' | X)$.
\textcolor{black}{ Following standard MCMC convergence arguments,} the induced Markov chain converges to the stationary distribution $p(X | \phi(s(\tilde{X}'), s(x')) < \epsilon)$. In general, choosing smaller values of $\epsilon$ will improve the approximation of our actual target $p(X | X'=x')$. Algorithm \ref{alg:mcmc} illustrates the whole procedure. If the 
approximation level required entails dealing with a very small $\epsilon$, the acceptance rate would drop, resulting again in poor mixing. A possibility to improve this consists of building a Markov chain in an augmented state-space $(x, \epsilon)$ \citep{bortot2007inference}. Values of $x$ simulated using large values of $\epsilon$ are less reliable but the transition to such values can improve mixing.

\begin{algorithm}[h!] 
\caption{MCMC-ABC sampler for $p(X | X' = x')$}
\label{alg:mcmc}
\begin{algorithmic}
\State {\bf Input:} Instance $x'$, distribution $p(X' | X )$, prior $p(X)$, transition density $q(x \rightarrow \tilde{x})$ tolerance $\epsilon$, set of statistics $s$ and distance $\phi$.
\State {\bf Output:} Samples approximately distributed according to $p(X | X' = x')$.
\State Initialize $x_0$, $i=0$.
\State {\bf Repeat \textcolor{black}{until convergence}:}
\State Propose $\tilde{x}$ according to transition distribution $q(x_i \rightarrow \tilde{x})$.
\State Sample $\tilde{x}' \sim p(X' | X= \tilde{x})$.
\State Compute
\begin{equation*}
        \alpha = \min \left \lbrace 1, \frac{p(\tilde{x}) q(\tilde{x} \rightarrow x_i)}{p(x_i) q(x_i \rightarrow \tilde{x})} \mathbb{I}\left[ \phi(s(\tilde{x}'), s(x') ) < \epsilon \right]\right \rbrace
    \end{equation*}
\State With probability $\alpha$ set $x_{i+1} = \tilde{x}$, otherwise $x_{i+1} = x_i$.
\State Set $i = i+1$.
\State \textbf{End Repeat}
\end{algorithmic}
\end{algorithm}

\noindent To complete the specification, we still need to be able to sample from or evaluate $p(X)$ and sample from $p(X' \vert X)$.

\textcolor{black}{
\subsubsection*{\textbf{Sampling from $p(X)$ and $p(X' \vert X)$.}}
Estimating $p(X)$ is standard using training data, 
which is untainted by assumption. We just need a density estimation technique. 
For example, approximately sampling from this distribution could be done via bootstrapping from the training data. In addition, other techniques such as generative adversarial networks \citep{NIPS2014_5423}, energy-based models \citep{grathwohl2019your} or \textcolor{black}{ mixture models \citep{gamma}} could be used. For high-dimensional $X$ this sampling might be complex,  partly motivating our  approach in Section \ref{sec:diff}}.

On the other hand, sampling from $p(X' | X)$ entails strategic thinking
as we need to model how the adversary would modify the originating instance $X$. Any attacking model in the literature, {\color{black} 
 including the ones sketched in Section \ref{attacks}},
could be used,  making our framework applicability very general. However, we propose a formal Bayesian decision-theoretic argument to produce samples from $p(X' | X)$, 
employing the ARA methodology to account explicitly for the uncertainty about the adversary's behavior.
In particular, we identify two sources of uncertainty that are relevant for 
 adversarial purposes: a) adversarial attacks might not be deterministic as, 
e.g., an attacker might choose to randomize between multiple different data transformations
and, thus, we should consider {\em aleatory} uncertainty when modeling attacks; b) another source of uncertainty ({\em epistemic}) stems from our lack of knowledge about the adversary. Herein, we show how the ARA methodology can be used to model both types of uncertainties.

With no loss of generality, assume that, out of the relevant 
$k$ classes, the attacker  deems as interesting the first $l$  (call them 
 {\em bad}), the other ones being irrelevant to him ({\em good}): he is interested in modifying data associated with instances belonging to the \textcolor{black}{ bad classes to make $C$ believe that they belong to the good ones}. As an example, consider a fraudster who may commit $l\,\,$ types of fraud. He crafts the corresponding $x$ to $x'$ to make the classifier think that she has received a legitimate instance from  class $y>l$. 
As we only consider integrity violations (that is, {\color{black} the adversary has 
just control over bad instances)} we use the decomposition 
\begin{eqnarray*}
    p(x' | x) &=& \sum_{y=1}^k p(x' | x, y) p(y | x) = \sum_{y=1}^l p(x' | x, y) p(y | x)  + \\ &+& \sum_{y=l+1}^k  \mathbb{I}(x'= x)p(y|x). 
\end{eqnarray*}
Sampling  from $p(y | x)$ is simple, as those probabilities can be estimated from training data.
Therefore, we can obtain samples from $p(x' | x)$ by first generating from $y \sim p(y | x)$ 
and, then, if $y > l$ return $x$ or sample $x' \sim p(x' | x, y)$, otherwise. 

\textcolor{black}{ We still need a procedure to sample from $p(x' | x, y)$. Again, any attack model could be used here, but we employ a Bayesian decision theoretic framework to model the adversary's decision problem when he has available an instance $x$ with label $y$,
 employing the ARA methodology.
Assume the attacker is an expected utility maximizer trying to fool $C$.} His utility function has the form $u_A (y_C ,y)$, when the classifier says $y_C$ and the actual label is $y$. 
The attacker should choose the feature modification maximizing his expected utility by making $C$ classify instances as most beneficial as possible to him. \textcolor{black}{
With no loss of generality, assume }  
the utility that $A$ derives from the classifier's decision has the  structure 
\begin{equation*}
  u_{A}(y_{C}, y) =\left\{
  \begin{array}{@{}ll@{}}
    0, & \text{if}\ y \leq l\ \text{and}\ y_C \leq l \\
    u_{A}^{y_C, y} > 0, & \text{if}\ y \leq l\ \text{and}\ y_C > l \\
    0 , & \text{if}\ y > l
  \end{array}\right.
\end{equation*} 
reflecting the fact that the attacker just obtains benefit when he makes the defender classify a bad instance as good.
By transforming instance $x$ with label $ y \in \lbrace 1,\dots,l \rbrace$ 
into $x'$, the attacker would get an expected utility 
\begin{equation}\label{HVN}
     \sum _{y_C=1}^k  u_A (y_C , y ) p_A (y_C | x' )
= \sum _{y_C=l+1 }^k  u_A^{y_C,y} \, p_A (y_C| x' ),
\end{equation}
with $p_A ( y_C | x' )$ describing the probability that $C$ classifies the observed instance $x'$ as $y_C$, from $A$'s perspective. 
%
{\color{black}
Thus, the adversary should craft instance $x$  with label $y$ into the attacked instance $x'(x, y)$ with
\begin{equation}\label{winston}
x' (x, y ) = \arg\max_{z}   
\sum _{y_C=l+1 }^k  u_A^{y_C,y} \, p_A (y_C| z ), 
\end{equation}
where the optimization is performed over the set of all possible modifications of instance $x$.
}

\textcolor{black}{However, since typically we shall not have access to the 
adversary to completely elicit his preferences and beliefs, 
 we model our epistemic uncertainty about $u_A^{y_C,y}$ and $p_A (y_C| z )$ in a Bayesian 
   way with, respectively, random utilities $U_A^{y_C,y}$ and random  probabilities $P_A (y_C| z )$ defined, with no loss of generality, over an appropriate common probability space 
$(\Omega,{\cal A},{\cal P})$ with atomic elements $\omega \in \Omega$ \citep{Chung}.}
This induces a distribution over the attacker's expected utility,
where the random expected utility for him would be
$
\sum _{y_C=l+1 }^k  U_A^{y_C, y, \omega } \, P_A^{y_C, \omega } ( z ).$
In turn, the random optimal attack is defined through 
$
X_{\omega }'(x, y)=\argmax_z \sum _{y_C=l+1 }^k  U_A^{y_C, y, \omega } \, P_A^{\omega} (y_C| z )$,  and we can make $p(x'|x, y)= {\cal P} (X_{\omega }' (x, y)= x')$. Defined this way, our model for $p(x'|x, y)$ properly accounts for the existing uncertainty about the adversary.
Note that, by construction, if we sample $u_A \sim U_A$ and $p_A (y_C| z ) \sim P_A (y_C| z )$ and solve (\ref{winston}),
%
%
$x'(x,y)$ would be distributed according to $p(x' | x, y)$.  As 
mentioned, this is the last ingredient needed to implement our approach and, thus, completes our \textit{adversarial modeling} framework  to 
adversarial classification. 

Observe that we have assumed that the attacker is an expected utility maximizer. Thus, if the data manipulation maximizing expected utility is unique, the generated attack \textcolor{black}{(given the attacker's utility and probability)} will be deterministic, in the sense that instance $(x,y)$ will always lead to the same manipulated $x'$.
Under the proposed framework, we could easily model a non-deterministic attacker that randomizes possible attacks.
One possibility would be to model the attacker as an agent that randomly selects an attack
in such a way that those with higher expected utility are more likely to be chosen. This would just require introducing an extra sampling step when simulating from $p(x' | x, y)$. \textcolor{black}{Thus it is possible to  not only integrate epistemic uncertainty into our attacker modeling strategy but also to account for aleatoric uncertainty.}

\subsubsection*{\textbf{Models for random utilities and random probabilities}}

\textcolor{black}{ We describe now general guidelines that facilitate the 
 specification of the remaining elements, that is models for 
random utilities $U_A$ and random probabilities $P_A(y_C| z )$.}
 Methodologically, these are prior assessment problems \citep{PRIOR}; as a general statement, instead of a blind choice of hyperparameters, we try to use all the structural information that is available. 
Importantly, we can adjust the variance of the proposed 
 prior assessments to reflect the reliability of the existing knowledge. Finally, we 
should submit the entire study to a sensitivity analysis, in particular with respect to the proposed priors \citep{EKIN2022,ruggeri}.
 


\textcolor{black}{Consider first the random utilities $U_A$. Recall that, without loss of generality, 
 we may scale utilities in our setup so that they have support $[0,1]$ 
 by just adopting an appropriate positive 
 affine transformation based on the axiomatic properties of utility functions \citep{french2000statistical}}. Among distributions with such support,
a convenient choice due to their flexibility has been using $U_{y_C,y}\sim \text{Beta} (\alpha _{y_C, y}, \beta _{y_C, y})$ distributions.
\textcolor{black}{In particular, if no further additional information 
is available, we could use, e.g., a uniform distribution \citep{yang}.} 
Rankings about perceived utilities are 
relatively easy to obtain and we just need to
introduce them as constraints and sample from the 
random utilities by rejection.
If eventually, further information about the likely values of the utilities is available, we may assess them through appropriate choices of $\alpha_{y_C, y}$ and $\beta_{y_C ,y}$, using standard expert judgment elicitation approaches as in \cite{Morris}. \textcolor{black}{ This information can be incorporated by appropriately choosing the mean of the Beta distribution. In addition, we can regulate its variance while keeping the mean constant in order to reflect different levels of knowledge.}

Modeling $P_A (y_C| x' )$ is more delicate. It 
entails strategic thinking as $C$ needs to model her opponent's beliefs about what classification she will adopt upon observing $x'$. Potentially, this leads to a hierarchy of recursive decision-making problems as the classifier  needs
{\em to think about what the attacker thinks about}... 
akin to level-k thinking in game theory \citep{stahl}, although we stop the entailed recursion at a level in which no more information is available when we introduce non-informative distributions over 
probabilities. We then solve recursively going down in the hierarchy by maximizing expected utility.
This procedure is illustrated in \cite{araCounter} in a simpler context. Here, we describe a level-2 modeling of $P_A (y_C| x' )$ as a distribution based
on $p (y_C | x' )$ with some uncertainty around it. 
For this, consider the set ${\cal X}_{x'}$ of reasonable origins given the received $x'$.
Since changing instances typically entails some cost for the adversary that increases with the number of
features crafted, a reasonable choice is to define 
${\cal X}_{x'}$ as the set of features $x$ such that $\lambda (x,x')<\rho$ for a certain metric $\lambda$
and threshold $\rho$. 
Next we consider an auxiliary distribution $p^* (x|x')$
over ${\cal X}_{x'}$; we adopt either 
a uniform distribution over ${\cal X}_{x'}$, or make $p^* (x|x' ) \propto [\lambda (x,x')]^{-1}$.
Then, we define  $\mu_{y_C}  = \sum _x p (y_C | x) p ^* (x|x' )$, where $p(y_C |x)$ would come from estimates based on untainted training data. 
\textcolor{black}{A Dirichlet distribution is used to model the behavior of the random vector encompassing all $k$ possible classifications $y_C$ given $x'$. The parameters will be chosen so that the mean for each $y_C$ will coincide with $\mu_{y_C}$. As a consequence of the properties of the Dirichlet distribution, the marginal distribution for each individual $y_C$ will be $P_A (y_C| x' ) \sim Beta (\alpha^{y_C}, \beta^{y_C})$ having mean $\mu_{y_C}$ and a variance $\text{var}_{y_C}$ properly chosen.
To reduce the computational cost, we could approximate $\mu_{y_C}$ through $\frac{1}{M}\sum_{n=1}^M  p(y_C| x_n)$, for a sample $\{ x_n \}_{n=1}^M$ from $p^*(x|x')$. } \\
{\color{black}
\subsubsection{\textbf{ Case. Robustifying classification algorithms in sentiment analysis. }}
We illustrate the proposed approach with the sentiment analysis problem from Example 1.
We test the performance of AB-ACRA as a defense mechanism against adversarial attacks, based on a 0-1 utility for the defender. 
\textcolor{black}{Our discussion focuses on the random forest classifier there used,
but we provide assessments for the other three classifiers in the example. 
As benchmark, recall that its accuracy over clean test data was $0.720 \pm 0.005$ (Table 1).}

To compare AB-ACRA with raw RF on tampered data, let us simulate attacks over the instances in the test set using \textit{attacker A}. For this, we solve problem \eqref{winston} for each test review, removing the uncertainty that is not present from the adversary's point of view, restricting to attacks that involve changing at most the value of two of the words for  each review. The utility that the attacker perceives when he makes the defender misclassify a review is 0.7. Finally, the adversary would have uncertainty about $p_A^{y_C}(x')$, as this quantity depends on the defender's decision. We test AB-ACRA against a worst case adversary who knows the actual value of $p(y_C \vert x_n)$ and estimates $p_A^c(x')$ through $\frac{1}{M} \sum_{n=1}^M p(y_C \vert x_n)$ for a sample $\lbrace x_n \rbrace_{n=1}^M$ from $p^*(x \vert x')$, with $M=40$. For $p^*(x \vert x')$ we use a uniform distribution on the set of all instances at distance 1 from the observed $x'$, using $\lambda (x, x') = \sum_{i=1}^{150} \vert x_i - x'_i \vert$ as distance. 
\textcolor{black}{As we are in a binary classification setting the uncertainty about $p_A^{y_C}(x')$ and the attacker's utility function from the defender's perspective is modeled through beta distributions centered at 
the attacker's utility and probability values, with variances chosen to guarantee that the distribution is concave in its support.} Otherwise, we would be believing that the attacker's utility and probability are peaked around 0 and 1 and low in between, which makes little sense in our context. For this, variances must be bounded from above by $\min \big \lbrace[\mu^2(1-\mu) ]/ (1+\mu), [\mu(1-\mu)^2] / (2-\mu) \big \rbrace$, were $\mu$ is the corresponding mean. The size of the variance will inform about the degree of knowledge the defender is assumed to have about the attacker. Reflecting a moderate lack of knowledge, we set the variance to be $10 \%$ of this upper bound. Thus, we are assuming a certain degree of knowledge about the adversary, as the expected values of the random utilities and probabilities coincide with the actual values used by the attacker. We later study how deviations from the assumed attacker behavior affect performance.

As summary statistic $s$ for AB-ACRA, we use
the 11 features (out of 150) having the highest permutation feature importance \citep{breiman2001random}.
Given the discrete nature of the covariates, the distance used in the ABC scheme is Hamming's. Figure \ref{fig:samples} compares the accuracy of AB-ACRA and RF for different sample sizes $N$ in Algorithm \ref{alg:gen_ara}, and  
tolerance $\epsilon = 2$. As we can see, AB-ACRA beats RF in tainted data with $N=5$. The accuracy saturates quickly as we increase the number of samples: 
good performance is achieved with a relatively small sample size. Figure \ref{fig:tolerance}
plots the accuracy of AB-ACRA against RF for different tolerance $\epsilon$ values: as this parameter decreases, accuracy increases, 
albeit at a higher computational cost. 
\begin{figure*}[t]
\begin{subfigure}{.5\textwidth}
  \centering
  \includegraphics[scale=0.6]{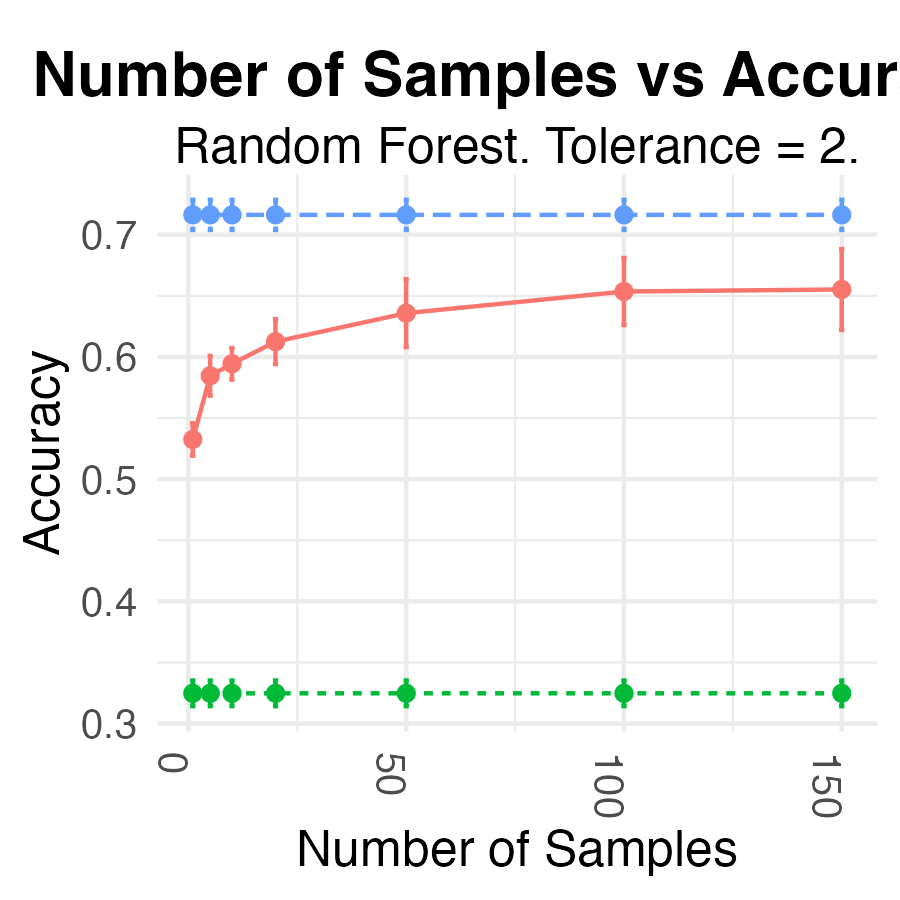}
  \caption{Experiment for a different number of samples.}
  \label{fig:samples}
\end{subfigure}%
\begin{subfigure}{.5\textwidth}
  \includegraphics[scale=0.6]{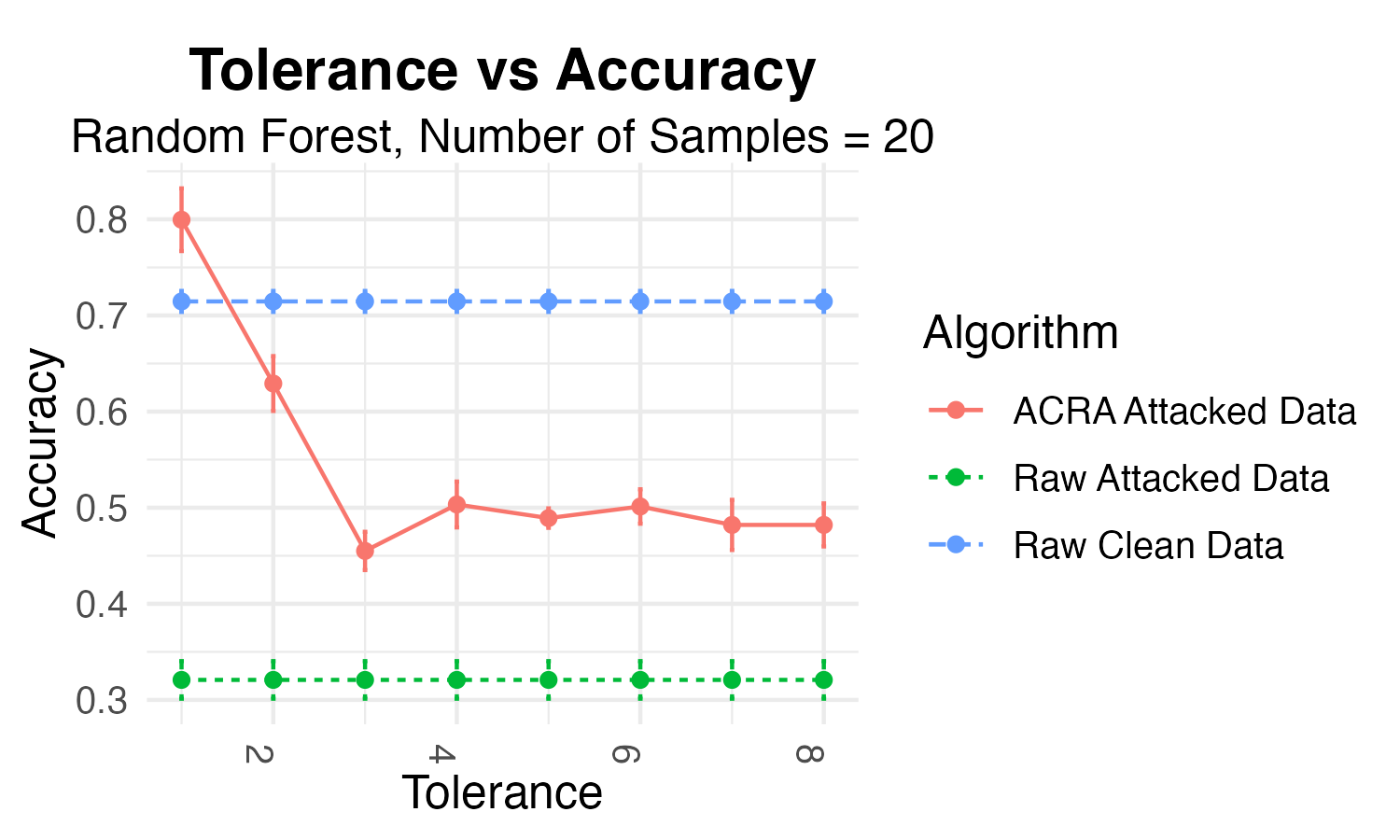}
  \caption{Experiment for different tolerance values.}
  \label{fig:tolerance}
\end{subfigure} 
\caption{Accuracy comparison RF vs AB-ACRA.}\label{fig:accSamplesTolerance}
\end{figure*}

Tables \ref{tab:rawVSprotectedA} and \ref{tab:rawVSprotectedB} show average accuracies of the four classifiers from Example 1 robustified during operations against tainted data using attackers A and B, respectively.  
As can be seen, our approach allows us to mitigate performance degradation by showcasing the benefits of explicitly modeling the attacker's behavior in adversarial environments. Interestingly, in most cases, the classifiers even perform better on attacked data than the raw algorithm on clean data. This regularizing effect was mentioned in  \cite{goodfellow2014explaining} for other algorithms and application areas.

\begin{table*}[t]

\caption{Accuracy comparison (with precision) 
 	of four classifiers with and without protection on clean and attacked data. Attacker A.}
\centering
\begin{tabular}[t]{cc|cc}
\toprule
\textbf{Classifier} & \textbf{ Clean data (Raw) } & \specialcell{ \textbf{ Attacked data (Raw)} \\ Attacker A} & \specialcell{\textbf{ Attacked data (AB-ACRA)} \\ Attacker A} \\
\midrule
Logistic Regression & $ 0.728 \pm 0.005 $ & $ 0.322 \pm 0.011 $ & $ 0.589 \pm 0.023 $\\
Naive Bayes & $ 0.722 \pm 0.004 $ & $ 0.333 \pm 0.009 $ & $ 0.968 \pm 0.008 $\\
Neural Network & $ 0.691 \pm 0.019 $ & $ 0.338 \pm 0.021$ & $0.761 \pm 0.030 $\\
Random Forest & $ 0.720 \pm 0.005 $ & $ 0.327 \pm 0.011$ & $ 0.837 \pm 0.014 $\\
\bottomrule
\end{tabular}
\label{tab:rawVSprotectedA}%
\end{table*}

\begin{table*}[t]

\caption{Accuracy comparison (with precision) 
 	of four classifiers with and without protection on clean and attacked data. Attacker B.}
\centering
\begin{tabular}[t]{cc|cc}
\toprule
\textbf{Classifier} & \textbf{ Clean data (Raw) } & \specialcell{ \textbf{ Attacked data (Raw)} \\ Attacker B} & \specialcell{\textbf{ Attacked data (AB-ACRA)} \\ Attacker B} \\
\midrule
Logistic Regression & $ 0.728 \pm 0.005 $ & $ 0.418 \pm 0.010 $ & $ 0.840 \pm 0.010 $\\
Naive Bayes & $ 0.722 \pm 0.004 $  & $ 0.405 \pm 0.009 $ & $ 0.880 \pm 0.027 $\\
Neural Network & $ 0.691 \pm 0.019 $ & $ 0.417 \pm 0.015 $ & $ 0.700 \pm 0.064 $\\
Random Forest & $ 0.720 \pm 0.005$  & $ 0.397 \pm 0.013 $ & $ 0.826 \pm 0.012 $\\
\bottomrule
\end{tabular}
\label{tab:rawVSprotectedB}%
\end{table*}


The previous experiments quantified the defender's uncertainty about the attacker's utility and probability through beta distributions centered around the values actually employed by the adversary. In realistic settings, these values are often unknown and estimates are instead used. Therefore, it is natural to explore how performance is impacted when using estimates that may be potentially far from the truth.
Our second batch of experiments tests the approach against an attacker whose utilities and probabilities differ from the baselines 
elicited by the defender. In particular, 
the attacker will 
deviate uniformly around the assumed probability and utility for each attack. The size of the deviation is constrained to be less than 25\% the assumed value: if we center our beta distribution for, e.g., the attacker's probability at value $\mu$, the attacker will deviate from the assumed behavior in the range $(0.75 \cdot \mu, 1.25 \cdot \mu)$, with the upper bound 
truncated to 1 if this value is exceeded.
Hence, in this experiment, our beta distributions will be centered around 
{\em wrong values}. 
We set the variance of the beta priors to be relatively high, $50 \%$ of the upper bound, and compare our approach with the CK one, in which the elements of the attacker are assumed to be known, and thus are point masses (on wrong values). 

Tables \ref{tab:ckVSnockA} and \ref{tab:ckVSnockB} show average accuracies 
of the four classifiers on attacked data without defense (col. 2), the standard 
CK defense (col. 3) and, finally, our AB-ACRA defense (col. 4). Note first that the overall performance drops with respect to the results in Tables \ref{tab:rawVSprotectedA} (col. 4) and \ref{tab:rawVSprotectedB} (col 4.): when the attacker deviates from the assumed behavior, the performance 
recovery of both AB-ACRA and CK defenses is worse. 
\textcolor{black}{ Importantly, these results suggest as well that when the adversary deviates from the common knowledge assumption, AB-ACRA is as accurate as the common knowledge approach for certain classifiers and more accurate for others. This reflects that accounting for the uncertainty over the attacker elements is indeed beneficial when the attacker deviates from the assumed behavior. This experiment showcases the increase in robustness due to modeling uncertainty in scenarios in which CK assumptions are not realistic.
}


\begin{table*}[t]

\caption{Accuracy comparison (with precision) 
 	of four classifiers on tainted data with no defense, CK defense, and AB-ACRA defense. Attacker A. \textcolor{black}{ Best 
  result in boldface}}
\centering
\begin{tabular}[t]{cccc}
\toprule
\textbf{Classifier} &  \specialcell{ \textbf{ Attacked data (Raw)} \\ Attacker A} & \specialcell{ \textbf{ Attacked data (CK)}\\ Attacker A} & \specialcell{ \textbf{ Attacked data (AB-ACRA)}\\ Attacker A} \\\midrule
Logistic Regression & $ 0.315 \pm 0.007 $ & $ 0.499 \pm 0.008 $ & $ \boldsymbol{0.513} \pm 0.008 $\\
Naive Bayes & $ 0.325 \pm 0.007 $ & $ 0.645 \pm 0.025 $ & $ \boldsymbol{0.665} \pm 0.024 $\\
Neural Network & $ 0.389 \pm 0.024 $ & $ 0.592 \pm 0.032 $ & $ \boldsymbol{0.638} \pm 0.030 $\\
Random Forest & $ 0.313 \pm 0.009 $ & $ \boldsymbol{0.720} \pm 0.013 $ & $ 0.710 \pm 0.017 $\\
\bottomrule
\end{tabular}
\label{tab:ckVSnockA}
\end{table*}

\begin{table*}[t]
\caption{Accuracy comparison (with precision) 
 	of four classifiers on tainted data with no defense, CK defense, and AB-ACRA defense. Attacker B. \textcolor{black}{ Best 
  result in boldface}}
\centering
\begin{tabular}[t]{cccc}
\toprule
\textbf{Classifier} &  \specialcell{ \textbf{ Attacked data (Raw)} \\ Attacker B} & \specialcell{ \textbf{ Attacked data (CK)}\\ Attacker B} & \specialcell{ \textbf{ Attacked data (AB-ACRA)}\\ Attacker B} \\\midrule
Logistic Regression & $ 0.412 \pm 0.004 $ & $ 0.713 \pm 0.008 $ & $ \boldsymbol{0.760} \pm 0.011 $\\
Naive Bayes & $ 0.406 \pm 0.008 $ & $ 0.783 \pm 0.039 $ & $ \boldsymbol{0.800} \pm 0.035 $\\
Neural Network & $ 0.437 \pm 0.014 $ & $ \boldsymbol{0.727} \pm 0.050 $ & $ 0.725 \pm 0.052 $\\
Random Forest & $ 0.402 \pm 0.005 $ & $ 0.779 \pm 0.011 $ & $ \boldsymbol{0.782} \pm 0.008 $\\
\bottomrule
\end{tabular}
\label{tab:ckVSnockB}
\end{table*}

}

\subsection{Robustifying classifiers during training}\label{sec:diff}

 \textcolor{black}{ Case 3.2.2 allowed us to illustrate the framework
  during operations in a 
relatively complex setup with a moderate number (150) of binary features and constraints on the 
 maximum (2) allowed number of changes.} However, the scheme presented may entail heavy computational costs in large-scale scenarios as sampling from $p(x \vert x')$ gets
 costly, even becoming infeasible 
computationally in high-dimensional 
domains. As an example, images are typically represented as matrices of size \emph{width}$\times$\emph{height} if in grayscale, or \emph{width}$\times$\emph{height}$\times$\emph{channels} if multiple channels of color are used. Even for baselines such as MNIST  in Example 2 with $28\times 28$ pixels, the feature number grows fast with the dimensions.

To overcome this computational bottleneck, 
\textcolor{black}{ we shall argue that some of the steps from the approach in Section \ref{sec:adv_aware_class} may be skipped
when dealing with differentiable classifiers}. 
By this, we understand classifiers whose \textcolor{black}{ structural form 
$p(y|\beta,x)$ is differentiable with respect to the $\beta$ parameters. 
 A particularly relevant case  
is }
\begin{eqnarray}\label{TGF}
&& p(y|\beta, x) = \mbox{softmax} (f_\beta (x))[y], \\
& \text{where} & \mbox{softmax}(x)[j] = \frac{\exp{x_j}}{\sum_{i=1}^k \exp{x_i} }, \nonumber
\end{eqnarray}
which covers a large class of models. For example,
if $f_\beta$ is linear in inputs, 
we recover multinomial regression (MR) 
\citep{mccullagh1989generalized}; if we take $f_\beta$ to be a sequence of linear transformations alternating non-linear activation functions, such as Rectified Linear Units (ReLU), we obtain a feed-forward neural network \citep{gallego}. 
These models are amenable to training through 
\emph{stochastic gradient descent} (SGD) \citep{bottou2010large}.  
     In particular, scalable optimization methods facilitate training deep neural networks
     (DNNs) 
     with large amounts of high-dimensional data, like images, since they 
      enable optimization using only mini-batches at each iteration. 

Importantly, instead of dealing with the attacker in the operational phase, as 
in Section 3.2, we shift to modifying the training phase to account for future adversarial perturbations. To enable so, we require the ability to draw posterior samples from the posterior distribution 
 $p(\beta|\mathcal{D})$.
With this paradigm shift, we avoid the 
expensive step of sampling from 
 $p(x|x')$, 
only requiring to do it from $p(x'|x)$ using gradient information from the defender model.
  \textcolor{black}{  Note that it is much easier to estimate } $p(x'|x)$ from an adversary,
just requiring an opponent model, than to estimate $p(x|x')$, which requires inverting such   opponent model. 
Obviously, since there is no gradient notion in every model,  we would need to resort to the approach in Section 3.2 in those cases.

For clarity, \textcolor{black}{  let us use $0-1$ utilities, although extensions to more general utilities follow a similar path}. We thus focus on implementing 
the decision rule 
\begin{align}\label{eq:decision_rule_5}
\argmax_{y_C} \iint p(y_C|x, \beta)p(x|x') p(\beta|\mathcal{D}) dx \, d\beta
\end{align}
in a general scalable and robust manner.

\subsubsection{\textbf{Protecting differentiable classifiers at training}}

Beyond usual adversarial robustness conditions such as 
(\ref{monday3}) and (\ref{tumorout}), 
our proposal will require the slightly stronger condition 
\begin{equation}\label{eq:cond1}
p(y,x|\beta) \approx p(y,x'|\beta).
\end{equation}
As we shall see, though this \textcolor{black}{ imposes some extra computational burden,
 it 
 motivates a scalable training scheme that 
improves the robustness of the classifier}.

 To start with, to compute $p(y,x|\beta)$,  we reinterpret 
its logits 
as in energy-based models \citep{grathwohl2019your}, \textcolor{black}{ 
leading to the expression 
$
p(y, x|\beta) = \dfrac{\exp \lbrace f_\beta(x)[y] \rbrace }{Z(\beta)},\,
$
where $Z(\beta)$ is the usually intractable normalizing constant.
Now, by factoring } the joint distribution $p(y,x|\beta)$ as $p(y|x,\beta)p(x|\beta)$,
then for a sample $x \sim \mathcal{D}$ and the corresponding adversarial perturbation $x' \sim p(x'|x)$, 
consider maximizing the objective function  ${\cal L} (\beta , x, y)$ with  %
\begin{eqnarray}\label{KKD}
    && \mathcal{L} (\beta, x, y) = \Big \lbrace \left[ \log p(y|x,\beta) + \log p(y|x',\beta) \right]  - \\
    &-&\left| f_\beta(x) - f_\beta(x') \right|  - \left| \log p(x|\beta) - \log p(x'|\beta) \right | \Big \rbrace. \nonumber
\end{eqnarray}
%
The first two terms in (\ref{KKD}) promote high predictive power for both $p(y|x,\beta)$ and $p(y|x',\beta)$. In turn, the third one encourages the logits of $x$ and \textcolor{black}{its corresponding perturbed sample} $x'$ to be similar, so that  $p(y|x,\beta) \approx p(y|x',\beta)$.
Finally, the last term acts as a new regularizer,  encouraging $p(x|\beta) \approx p(x'|\beta)$ and, together with the previous term, leads to condition (\ref{eq:cond1}). 
Interestingly, note that since in this third term we assess  the difference $\, \log p(x|\beta) - \log p(x'|\beta) \,$ of the originally intractable terms, the normalizing constant $Z(\beta)$ cancels out, 
rendering  tractable the analysis in (\ref{KKD}).

As a consequence, at the end of the training, by maximizing (\ref{KKD}) 
we would expect that $p(x|\beta) \approx p(x'|\beta)$.  Therefore, using Bayes formula, we would 
expect that $p(x|x') \approx p(x'|x)$. 
Then, considering that $p(y|x,\beta) \approx p(y|x',\beta)$, we  swap the original decision rule (\ref{eq:decision_rule_5}) by
\begin{equation}
\argmax_{y_C} \iint  p(y_C|x',\beta)p(x'|x)p(\beta|\mathcal{D})dx' d\beta .
\end{equation}
Note that since the observed input $x'$ might be tainted, it is not necessary to attack it via $p(x'|x)$ anymore and just suffices to use the test time decision rule, $\argmax_{y_C} p(y_C|x',\beta)$. 

To sum up, by promoting $p(x|\beta) \approx p(x'|\beta)$ during training, we 
learn a robust model for which starting in either $x$ or $x'$, it does not matter whether we use $p(x'|x)$ or $p(x|x')$ as we arrive at the same distribution, being much simpler to sample from $p(x'|x)$ 
  \textcolor{black}{  than from $p(x|x')$}, as explained
below. 
We emphasize that (\ref{KKD}) is more than just a mash-up of AT, ALP, and our new regularizer, with certain computational 
advantages. \textcolor{black}{In addition, (\ref{KKD}) can be seen as an augmented model likelihood, in which we expand the original model likelihood $p(y|x, \beta)$ with extra regularizers to improve adversarial robustness.} 

The next paragraphs apply the ARA methodology to add a layer of uncertainty over the previous terms with two objectives: i) enabling departure from
the above mentioned standard CK assumptions typical in AC; and ii) enhancing robustness and preventing overfitting. 
  Indeed, based on ARA, we 
acknowledge the two sources of uncertainty that motivated our interest in AC 
and bring in further realism to the proposed analysis: 
 the defender lacks full knowledge about the specific attack employed by her adversary and the latter usually does not have full knowledge of the model he desires to attack.

\textcolor{black}{ To address the first source, instead of performing an optimization to arrive at a single point as in e.g.\ (\ref{monday1})},
we replace SGD
 with an SG-MCMC sampler such as stochastic gradient
 Langevin dynamics (SGLD) \citep{welling2011bayesian} to sample from regions with high adversarial loss, thus being proportional to $\exp \lbrace - \log p(y|x,\beta) \rbrace$.
 This leads to iterations
 \begin{equation*}
x_{t+1} = x_t -\epsilon_t\, \mbox{sign} \nabla_x \log p(y|x_t,\beta) + \xi_t
\end{equation*}
with  $\xi_t \sim \mathcal{N}(0, 2\epsilon_t)$ and $t=1,\ldots,T$, where $\epsilon_t$ are step sizes that decay to zero following the usual  
\cite{robbins1951stochastic} convergence conditions. We also consider uncertainty over the hyperparameters $\epsilon_t$ (from a Gamma distribution, or better a re-scaled Beta, since too high or too low learning rates are futile) and the number $T$ of iterations  (from a Poisson). 
In addition, we can consider mixtures of attacks, for instance by sampling a Bernoulli random variable and, then, choosing the gradient corresponding to either FGSM or another 
attack from those in Section 2.2. 
Algorithm \ref{alg:large_attack} \textcolor{black}{  generates $K$ adversarial 
examples that take  into account the uncertainty
over the attacker's model and are amenable to large-scale settings.}

\begin{algorithm}[hbt] %
\caption{Large scale attack simulation}  
\label{alg:large_attack}
\begin{algorithmic}
\State {\bf Input:} Defender model $p(y|x,\beta)$, a set of $K$ particles $\lbrace \beta_i \rbrace_{i=1}^K$ and attacker model $p(x'|x)$ \textcolor{black}{(in this example, FGSM with multiple rounds)}. 
\State {\bf Output:} A set of adversarial examples $\{x_i\}_{i=1}^{K}$ from  attacker model.  
\State Sample $T \sim p(T)$
\State Sample $\epsilon \sim p(\epsilon)$
\For{each attack iteration $t$ from $1$ to $T$}
\State $x_{i,t+1} = x_{i,t} -\epsilon_t\, \mbox{sign} \nabla \log p(y|x_{i,t},\beta) + \mathcal{N}(0, 2\epsilon I)$
\EndFor
\State \textbf{Return} $x_i = x_{i,T}$
\end{algorithmic}
\end{algorithm}
\textcolor{black}{ Concerning the second source of uncertainty,} since the attacker may not know the actual $p (y|x,\beta)$, our  model
for his behavior takes into account the uncertainty 
over $\beta$. 
A first possibility considers an augmented model $p(y|x, \beta, \gamma)$
with $\gamma \sim \mathcal{B}er(p)$. Then,
for example, if $\gamma = 0$, $p(y|x, \beta, 0)$
may be given by MR, whereas if $\gamma = 1$, $p(y|x,\beta ,1)$ 
is a neural network. This would reflect the lack of information that the attacker has about the architecture he is targeting.
The case can be straightforwardly implemented as an ensemble model \citep{hastie2009ensemble},
performing simulated attacks over it.
Alternatively, $\beta$ may have continuous support. In the case of a NN, this would reflect that the attacker has uncertainty over the parameter values \citep{muller}. This 
can be implemented using scalable Bayesian approaches in deep models, such as SG-MCMC schema
\citep{ma2015complete}. To this end, we propose the defender model
to be trained using SGLD, obtaining posterior samples via the iteration
$\beta_{t+1} = \beta_t \textcolor{black}{+} \eta \nabla_\beta (\mathcal{L} (\beta_t, x, y) +\log p(\beta) )+ \mathcal{N}(0, 2\eta I),$ with objective $\mathcal{L} (\beta, x, y)$
as in (\ref{KKD})
  and sampling $x'$  using $p(x'|x)$ as in the previous paragraph. \textcolor{black}{Note that $p(\beta)$ is the prior placed over the model's parameters, which in the case of differentiable classifiers such as deep neural networks is typically a zero-centered Gaussian with scale in the order of 0.01.}
Algorithm \ref{alg:large_ara} employs the previous perturbations
\textcolor{black}{ to robustly train the classifier using ARA principles}.
\begin{algorithm}[hbt] %
\caption{Large scale ARA training}  
\label{alg:large_ara}
\begin{algorithmic}
\State {\bf Input:} Defender model $p(y|x,\beta)$ and attacker model $p(x'|x)$. 
\State {\bf Output:} A set of $K$ particles $\{\beta_i\}_{i=1}^{K}$ approximating the defender posterior 
 learned using ARA training.  
\For{each training iteration $t$}
\State sample $x_1, \ldots,  x_K \sim p(x' | x)$ using Alg. \ref{alg:large_attack}
\State $\beta_{i,t+1} = \beta_{i,t} + \epsilon \nabla (\mathcal{L}(\beta_{i,t}, x_i,y) +\log p(\beta_{i,t}))+ \mathcal{N}(0, 2\epsilon I)$ for each $i$ (SGLD)
\EndFor
\State \textbf{return} $\beta_i = \beta_{i,T}$
\end{algorithmic}
\end{algorithm}

 \textcolor{black}{Observe that to sample}
from the  posterior $p(\beta| \mathcal{D}_{att})$,
where 
$\mathcal{D}_{att}$ designates the attacked dataset, the Langevin SG-MCMC sampler can be written as
$$
\beta_{t+1} = \beta_{t} \textcolor{black}{+} \epsilon \nabla \log p(\beta | \mathcal{D}_{att})  + \mathcal{N}(0, 2\epsilon I).
$$
Since $\log p(\beta | \mathcal{D}_{att}) = \log p(\mathcal{D}_{att} | \beta ) \textcolor{black}{+} \log p(\beta) \textcolor{black}{-} \log p(\mathcal{D}_{att})$ and, taking gradients wrt to $\beta$, we have 
$\nabla \log p(\beta | \mathcal{D}_{att}) = \nabla \log p(\mathcal{D}_{att} | \beta ) \textcolor{black}{+} \nabla \log p(\beta)$.  \textcolor{black}{The previous sampler uses the 
log-likelihood under the attacked data. To further robustify it, we can replace that likelihood with the objective (\ref{KKD}), that also includes the likelihood over the clean samples
and the regularization term for improved adversarial robustness, in the sense of condition (\ref{eq:cond1}). Assume we work with a set of $K$ samples $\lbrace \beta_i \rbrace_{i=1}^K$ from the defender posterior, then an unbiased estimator of the gradient is obtained by sampling a minibatch $\lbrace x_i \rbrace$ of attacked points  using Algorithm \ref{alg:large_attack}, leading to the sampler
$$
\beta_{i, t+1} = \beta_{i, t} + \epsilon \nabla (\mathcal{L}(\beta_{i,t}, x_i,y) + \log p(\beta_{i,t}))   + \mathcal{N}(0, 2\epsilon I).
$$
}

 \textcolor{black}{ Finally, Algorithm \ref{alg:gen_ara_diff}, to be compared with Algorithm
 1, integrates and aggregates 
 the general procedure to robustify a classifier in a scalable manner.}

\begin{algorithm*}[htb] %
\caption{ARA procedure for Adversarial Training of Differentiable Classifiers}  \label{alg:gen_ara_diff} 
\begin{algorithmic}
\State {\bf Input:} training data $\mathcal{D}$, prior $p(\beta)$.
\State {\bf Output:} A classification decision $y_C^*(x')$.
\Train
\State Use Algorithm \ref{alg:large_ara} to obtain an approximation of $p(\beta | \mathcal{D})$ (robustified posterior).
\EndTrain
\Operation
\State Read instance $x'$
\State Find  
{\small 
$y_C^* (x')= \argmax_{y_C}\sum_{y=1}^k   u (y_C , y )  \int  p (y | x', \beta ) p(\beta|\mathcal{D})d\beta  
$ } 
\EndOperation
\State \textbf{return} $y_C^* (x')$
\end{algorithmic}
\end{algorithm*}

\subsubsection{\textbf{Case. Robustifying deep neural networks in computer vision.}}
\textcolor{black}{We apply the proposed approach to two mainstream datasets in computer vision,
MNIST and CIFAR-10, showcasing its benefits via experiments.} 

\textcolor{black}{In the first experiment, MNIST, 
the defender aims to classify digits (from 0 to 9) in presence of
adversarial attacks, recall Example 2. 
The underlying classifier is a two-layer feed-forward neural network with ReLU activations and a final softmax layer to get the predictions over the 10 classes \citep{gallego}. 
The net is trained using SGD with momentum $0.5$ for 5 epochs, learning rate of $0.01$, and batch size of 32. The training set corresponds to 50000 digit images
and results are reported over a 10000 digit test set. We use 
both of the uncertainties (attacker and defender) mentioned above, except that we do not adopt mixtures of different attacks or different models, focusing on a single-attacker setup.}

\begin{figure*}[htb]
\centering
\begin{subfigure}{.3\textwidth}
  \centering
  \includegraphics[width=.8\linewidth]{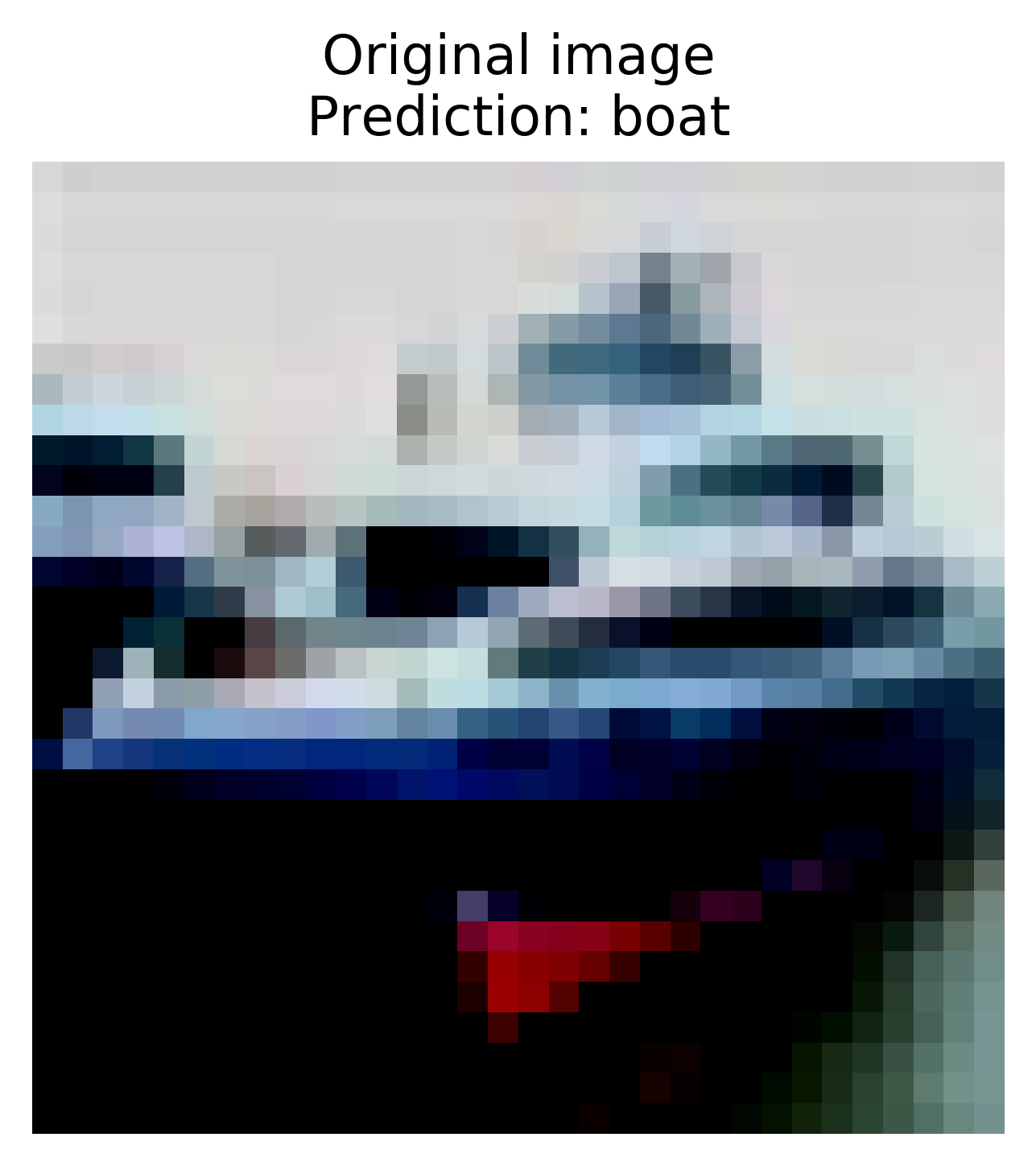}
  \caption{Original image \\(no attack)} 
  \label{fig:cifar_sample_1}
\end{subfigure}%
\begin{subfigure}{.3\textwidth}
  \centering
  \includegraphics[width=.8\linewidth]{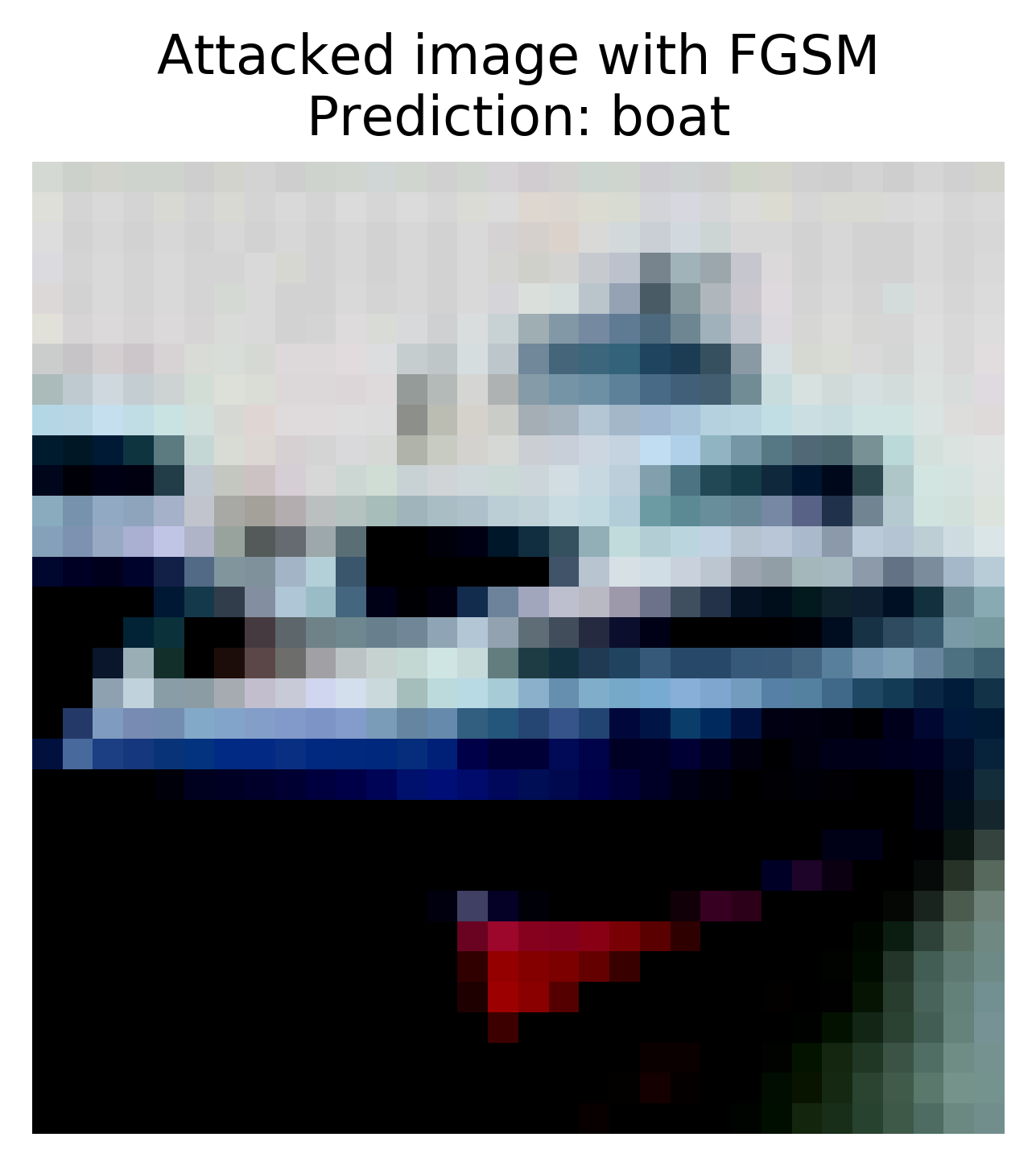}
  \caption{Perturbed image \\under FGSM intensity 0.01}
  \label{fig:cifar_sample_2}
\end{subfigure}
\begin{subfigure}{.3\textwidth}
  \centering
  \includegraphics[width=.8\linewidth]{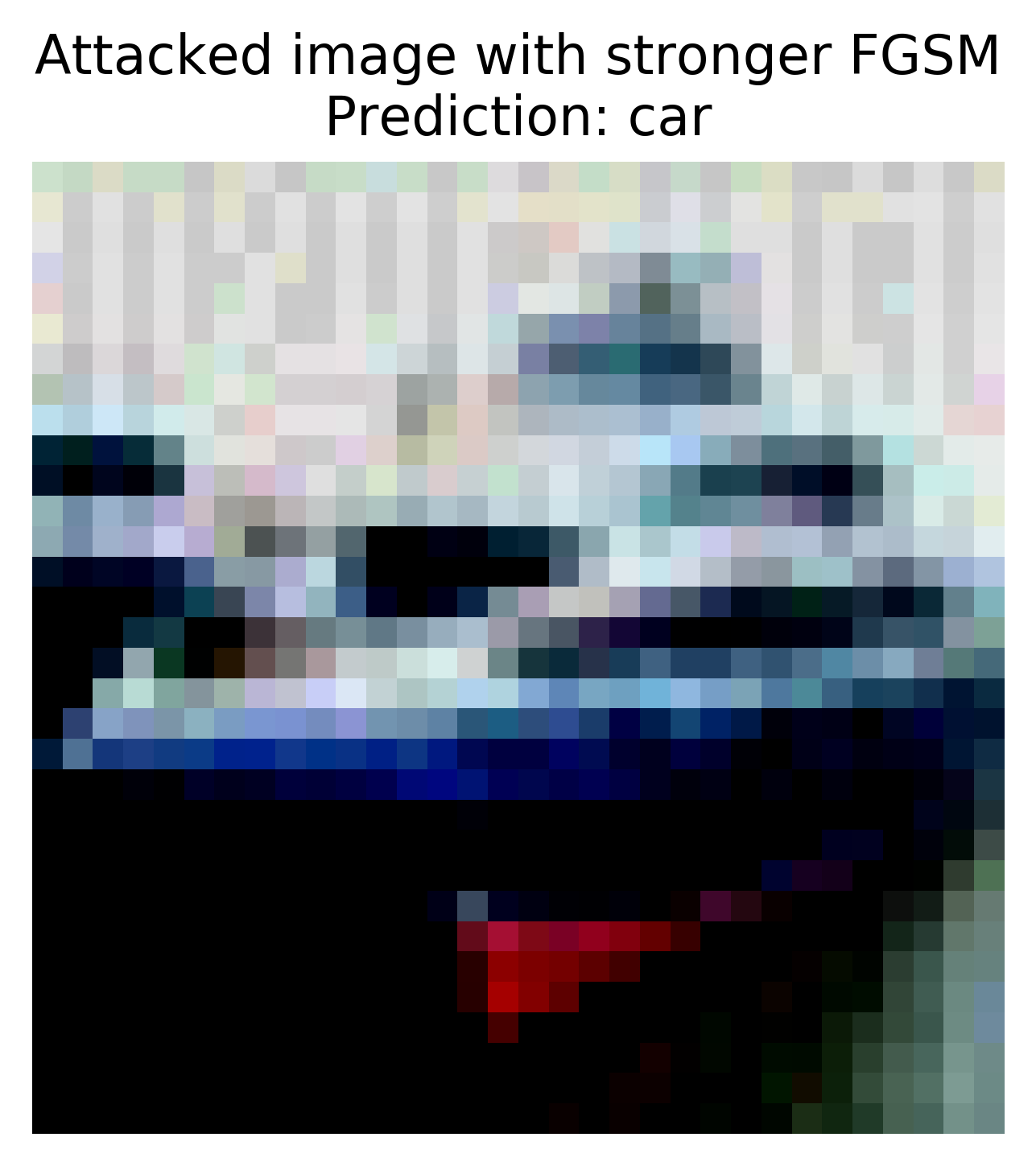}
  \caption{Perturbed image \\under FGSM intensity 0.04}
  \label{fig:cifar_sample_3}
\end{subfigure}
\caption{\textcolor{black}{A figure sample from CIFAR non-attacked (a) and under two 
increasingly intensive attacks (b and c).} }
\label{VICTOR}
\end{figure*}

\begin{figure*}[htb]
\centering
\begin{subfigure}{.5\textwidth}
  \centering
  \includegraphics[width=.99\linewidth]{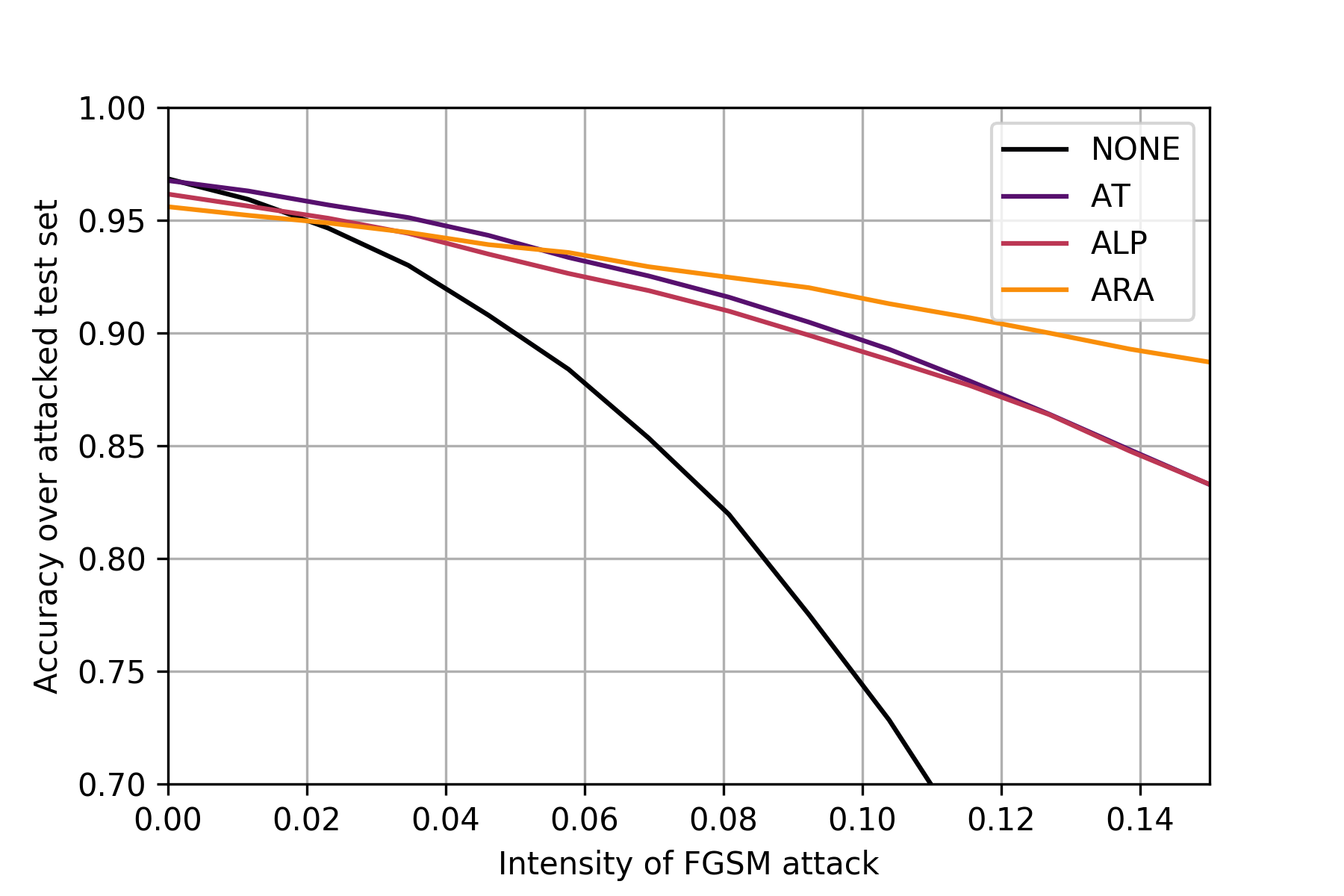}
  \caption{FGSM attack.}
  \label{fig:sub1a}
\end{subfigure}%
\begin{subfigure}{.5\textwidth}
  \centering
  \includegraphics[width=.99\linewidth]{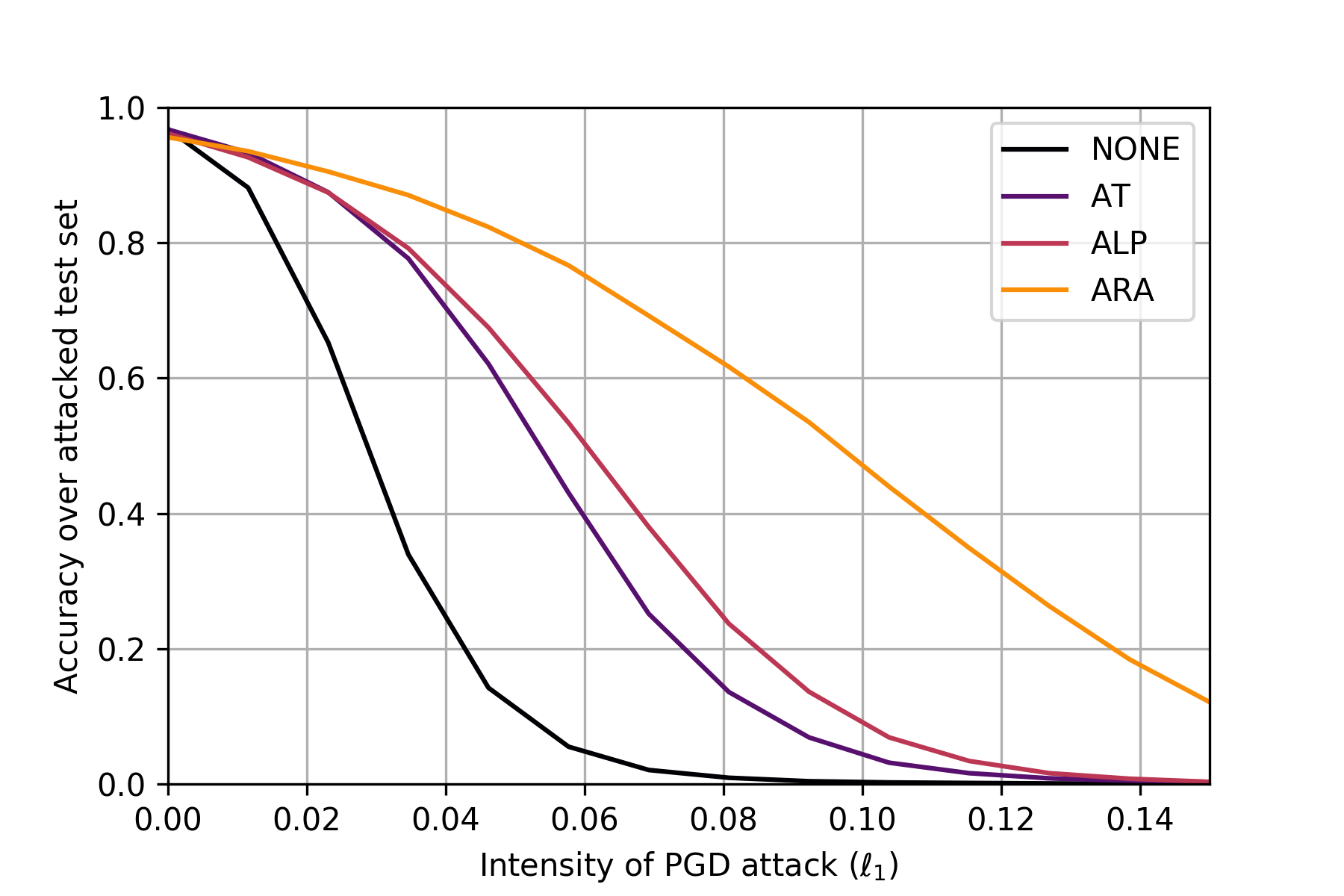}
  \caption{PGD attack under $\ell_1$ norm.}
  \label{fig:sub2a}
\end{subfigure}
\caption{Robustness of deep network for MNIST under three defense mechanisms
(ARA, AT, ALP) and two attacks (FGSM, PGD).} \label{fig:comparison_mnist}
\end{figure*}

\textcolor{black}{ 
Figure \ref{fig:comparison_mnist} plots the \emph{security evaluation curves} \citep{BIGGIO2018317}  
for three different defenses (AT, ALP, and our ARA-based proposal) and an undefended model (NONE in the legends) under the MNIST dataset, using two attacks at test time: FGSM (left) and 
PGD (right). Such curves depict the accuracy of the defender model ($y$-axis), under different attack intensities $\epsilon$ ($x$-axis),
 with larger attack intensities implying more powerful attacks, as Figure \ref{VICTOR}
 exemplifies. Note that whereas the first FGSM attack fails to change the predicted label, the stronger FGSM attack successfully makes the network predict an incorrect label.} 

  \textcolor{black}{ Inspection of Figure \ref{fig:comparison_mnist} immediately reveals noteworthy patterns.
  At low attack intensities, all four defenses perform comparably, although AML defenses appear to yield marginally lower accuracies. 
 However, as attack intensity increases, the performance of the undefended technique rapidly deteriorates illustrating the relevance of AML defenses. The undefended model's accuracy on the untainted data is 98\% and quickly degrades to 75\% under FGSM attack at an intensity of 0.1. The three robustified approaches mitigate this degradation, with varying degrees of success. The AT and ALP defenses perform comparably under FGSM; however, the AT defense degrades quicker under the PGD attack. The ARA approach generally appears more robust than the AT and ALP defenses at much higher intensities for both attacks. For example, a 0.1 FGSM-attack intensity induces AT and ALP accuracies of 90\% but an ARA accuracy of 92\%; this difference increases further with FGSM-attack intensity. The disparities are even more pronounced under the PGD attack. Although such attacks degrade all defenses, higher attack intensities are required to affect the ARA defense than the AT and ALP defenses.
 Thus, Figure \ref{fig:comparison_mnist} suggests that the uncertainties provided by the ARA training method substantially improve the robustness of the neural network under two different attacks with that provided by  state-of-the-art AT and
 ALP defenses.  Note also that robustifying the model against attacks is essential as its performance rapidly deteriorates.} 
 
We also compute the \emph{energy gap}
$\Delta E :=$ \\ $ \mathbb{E}_{x \sim \mathcal{D}}  \left[ - \log p(x) \right] - \mathbb{E}_{x' \sim \mathcal{D'}}  \left[ - \log p(x') \right]$ (using Eqs. (6) and (7) in 
 \cite{grathwohl2019your})
for a given test set $\mathcal{D}$ and its attacked counterpart $\mathcal{D'}$
under the PGD attack. This serves as a proxy to measure the degree of fulfillment of our 
enabling assumption (\ref{eq:cond1}).
We obtain that $\Delta E_{None} = 2.204, \Delta E_{AT} = 1.763$,
and $\Delta E_{ARA} = 0.070$.
The ARA version thus reduces the gap with respect to their counterparts,
getting closer to the desired adversarial assumption that a robust model
should fulfill  ($p(x) \approx p(x')$), 
\textcolor{black}{ having a clear
 regularization effect which 
contributes to our enhanced ARA approach.}

\begin{figure*}[htb]
\centering
\begin{subfigure}{.5\textwidth}
  \centering
  \includegraphics[width=.99\linewidth]{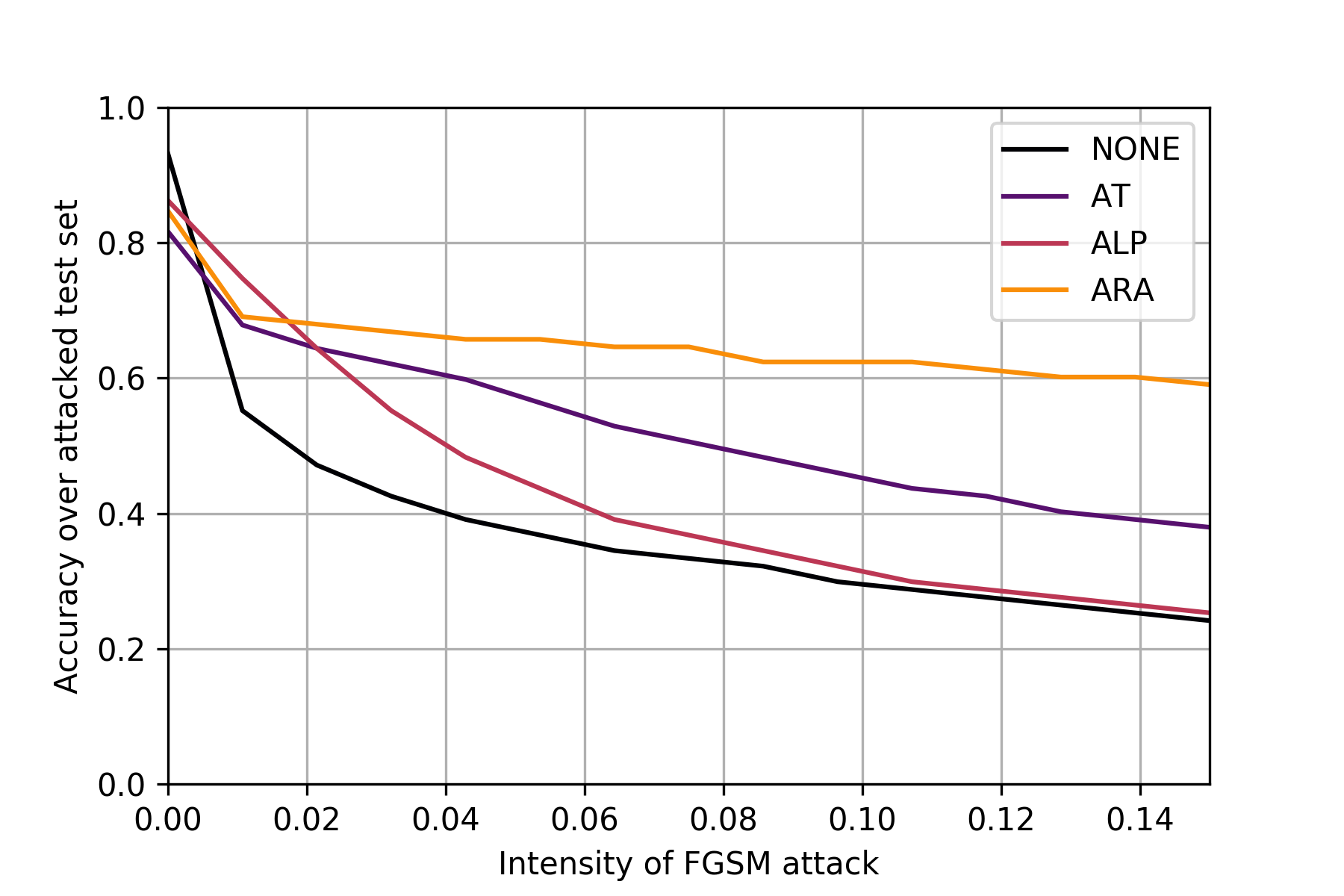}
  \caption{FGSM attack.}
  \label{fig:sub1b}
\end{subfigure}%
\begin{subfigure}{.5\textwidth}
  \centering
  \includegraphics[width=.99\linewidth]{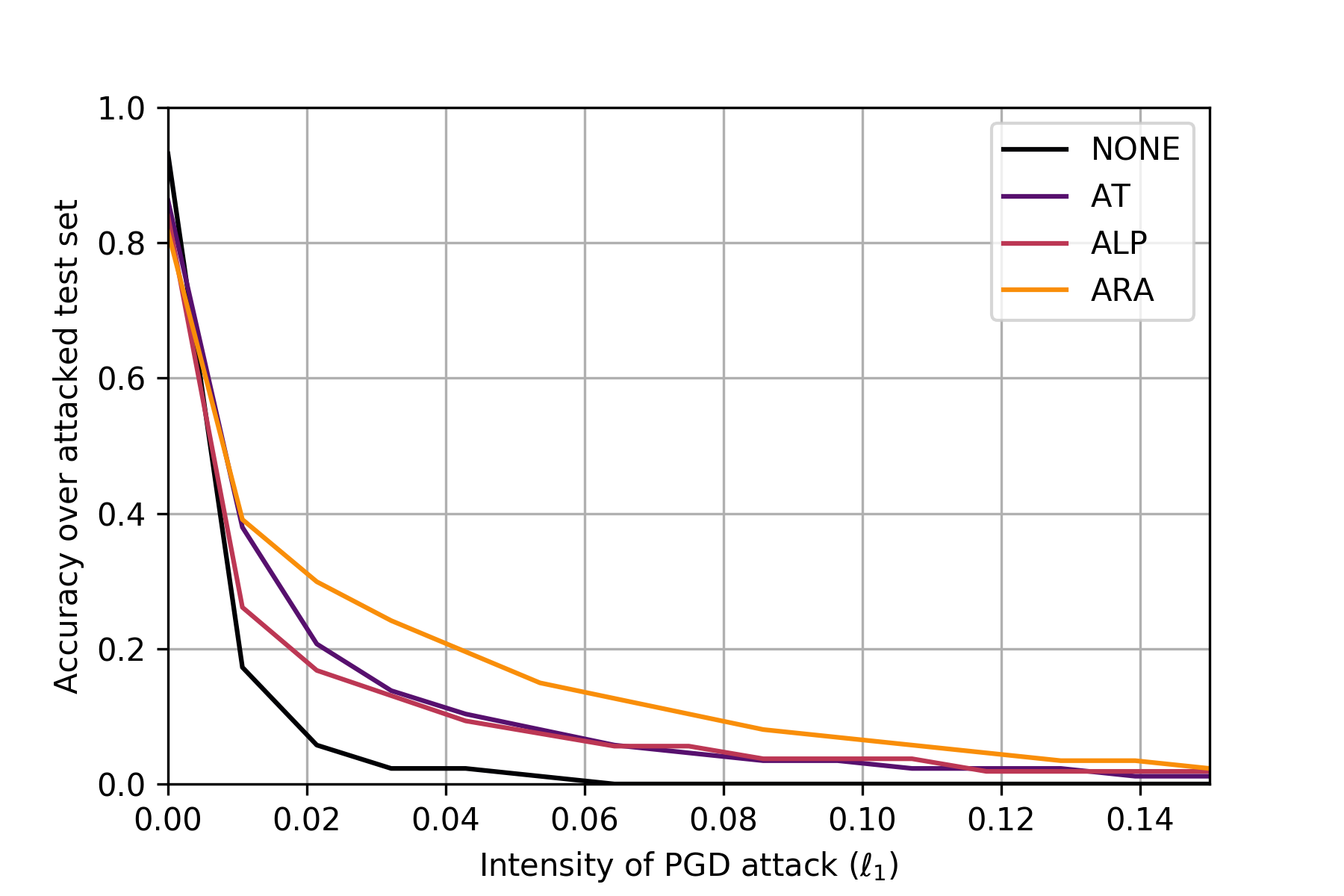}
  \caption{PGD attack under $\ell_1$ norm.}
  \label{fig:sub2b}
\end{subfigure}
\caption{Robustness of a deep network for CIFAR-10 under three different defense mechanisms (ARA, AT, ALP) and two attacks (FGSM, PGD).}\label{fig:comparison_cifar}
\end{figure*}

\textcolor{black}{
Figure \ref{fig:comparison_cifar} displays the same setting, analysing the 
protection from attacks over the classic CIFAR-10 dataset, which includes 
60000 $32\times 32$ color images (thus, of dimension 3072)
in 10 classes.
A much more complex architecture, a deep residual network consisting 
of 18 layers \citep{he2016deep,gallego}, is required 
for classification purposes. A similar picture emerges 
justifying the need for robustifying classifiers against 
adversarial attacks and our ARA improving upon AT and ALP 
defenses. Observe though, Fig. 5 b, that PGD attacks stress 
considerably the defenses in  such a complex problem.}

\section{A pipeline for adversarial classification}

We end up with the third major issue which refers to global frameworks for AML. 
\textcolor{black}{Due to its historical significance, we start by sketching \cite{adversarialClassification2004} pioneering approach to enhance classification algorithms when an attacker is present, 
adapting it to our notation (although they only focus on the utility-sensitive naive Bayes 
  classifier for binary problems).}
The authors view the problem as a game between $C$ and $A$,
using the following forward myopic approach.

\begin{enumerate}
\item \textit{C} $\,\,$ first assumes that data is untainted and computes her optimal classifier through 
\[ 
 \arg\max_{y_C} \sum_{i=1}^k
u (y_c , y_i ) p  (y_i | x ), 
\] 
where $p(y_i | x)$ is inferred using training data, clean by assumption.
\item Then, assuming that \textit{A} has complete information about the classifier's elements (a  CK 
assumption) and that $C$ is not aware of the 
   attacker's presence, the authors 
compute $A$'s optimal attack.  To that end, they propose solving an integer programming  problem, 
reflecting  the fact that the adversary tries to minimize the cost of modifying an instance, provided that such modification induces the change/s in the classification decision that $A$ is interested in.

%
%

\item Subsequently, the classifier, assuming that $A$ implements the previous attack (again a CK assumption) and that the training data is untainted, deploys her optimal classifier against it, by choosing $y_C$ maximizing $\sum_{i=1}^k u(y_C, y_i) p(y_i |x')$, her posterior expected utility given that she observes the possibly modified instance $x'$. This corresponds to  optimizing 
\begin{eqnarray*}\label{dalviCK}
\sum_{i=1}^k 
u(y_C, y_i) p(x' |y_i) p(y_i) .
\end{eqnarray*}
Estimating all these elements is straightforward, except for $p_C(x' \vert y_i)$. Again, appealing to a CK assumption, the authors assume that $C$, who knows all 
of $A$'s elements, can exactly solve 
the adversary problem from step two, and thus compute $x' = a(x, y_i)$, the attack deployed by the adversary when he receives instance $x$ with label $y_i$. Thus,
\begin{eqnarray*} 
p(x' |y_i) = \sum_{x \in \mathcal{X}'} p (x \vert y_i) p (x' \vert x, y_i)
\end{eqnarray*}
where $\mathcal{X}'$ is the set of possible instances possibly leading to the observed one after an attack and $p(x' \vert x, y_i) = 1$ if $a(x, y_i) = x'$ and 0 otherwise.
%
%
\end{enumerate}
The procedure could continue for more stages.
However, the authors consider sufficient to use these three.
As presented (and actually Dalvi et al stress 
in their paper), very strong common knowledge assumptions are made: all parameters of both players are known to each other. Although standard in game theory, such  an assumption is unrealistic in the security scenarios  
typical of AC yet has pervaded most of the later
literature in the field.


{\color{black}
Throughout the paper, we have emphasized an alternative view of AC that can be condensed into a general pipeline that consists of three main activities: gathering intelligence, forecasting likely attacks, and protecting classification algorithms. A first attempt to organize AML research within such three stages framework is due to \cite{BIGGIO2018317}. However, their framework relies on unrealistic CK assumptions.
As such, the authors propose to forecast likely attacks as solutions to certain constrained optimization problems, thus entailing deterministic attacks.

In line with our discussion in Sections 2 and 3, we provide a probabilistic 
version that mitigates these hypotheses. As argued in previous sections, attacks are probabilistic in order to reflect the lack of knowledge about the attacker's problem.  The steps are then:

\begin{enumerate}
    \item \textbf{Gathering intelligence.} The goal of this stage is to model the attacker's problem. This requires assessing attacker \textit{goals}, \textit{knowledge}, and \textit{capabilities}. As a result, a model for how the adversary manipulates an instance with covariates $x$ and label $y$ is constructed.

    One possibility to construct such a model is to use a normative decision-theoretic perspective, where the adversary is assumed to behave as a rational agent choosing data manipulations to maximize expected utility.
    \begin{equation} \label{adv}
        x' (x, y) = \argmax_{z} \sum_{i=1}^k u_A(y_C, y_i) p_A(y_C \vert z = a(x)),
    \end{equation}
    where $p_A(y_C \vert z = a(x))$ models the adversary's belief about the defender's decision upon observing the manipulated instance $z = a(x)$. 
    
    \item \textbf{Forecasting likely attacks.} The goal of this stage is to produce an \textit{attacking model} that incorporates not only the information gathered in step 1 but also the uncertainty that we have about the adversary's elements. Within our framework, an attacking model is assimilated with a probability distribution over attacked covariates $p(x' \vert x)$ given unattacked ones. Evaluating such a model is generally unfeasible but, as presented in Section 3, sampling from it is sufficient. 

    To sample from the attacking model, first, observe that $p(x' | x) = 
    \sum_{i=1}^k p(x' | x, y_i) p(y_i | x)$. Sampling from $p(y_i | x)$ is standard. Sampling from $p(x' | x, y_i)$ is more complex, as we lack information about how the adversary will modify an instance with covariates $x$ and true label $y_i$. In Dalvi's framework, as
     sketched above, CK was assumed, and as a consequence, $p(x' | x, y_i)$ was a point mass on the optimal adversarial modification of instance $(x, y_i)$. 
    Instead,  we propose modeling our uncertainty about the adversary placing priors on the utilities and probabilities in \eqref{adv}. This induces a distribution over the Attacker's optimal attack defined through
    \begin{equation*}
        X_\omega' (x, y) = \argmax_z \sum_{i=1}^k U^\omega(y_C, y_i) P^\omega_A(y_C \vert z ) 
    \end{equation*}
    where $U_A$ and $P_A^{y_D}$ are random utilities and probabilities defined over an appropriate common probability space 
    $(\Omega,{\cal A},{\cal P})$ with atomic elements $\omega \in \Omega$. 
    Then, by construction, $p(x' | x, y) = \mathcal{P}(X_\omega' (x, y) = x')$.
    Sampling from this distribution requires sampling from the random utilities and probabilities and computing the optimal attack conditioned on those samples.
     
    \item \textbf{Protecting classifiers.} Once with a reasonable attacking model, the last step is to protect the classifier against it. As we have seen, this can be done either at operations, modifying the way decisions about the label of a new instance are made; or at training, changing the way it is done to anticipate the future presence of an adversary. 

    During operations, an adversary-aware classifier will determine the label of a possibly modified instance $x'$ maximizing
    \begin{equation*}
        \sum_{i=1}^k  \int_{\mathcal{X}_{x'}} u(y_C, y_i)p(y_i | x) p(x | x') \dd x ,
    \end{equation*}
    where $x$ are the covariates of the unknown originating instance, which are marginalized out. In order to solve this problem, samples from $p(x | x')$ need to be generated, using steps 1 and 2 of the pipeline to sample from $p(x' | x)$ and e.g.\ the AB-ACRA approach proposed to generate from $p(x | x')$.

    Finally, protecting during training requires modifying how inference about the parameters of a given model is made, to guarantee robustness to adversarial manipulations, expressed through condition \eqref{eq:cond1} incorporated into (12).
    
\end{enumerate}

A natural question  is how to deal with the case in which the attacker modifies its behavior as a response to an implemented defense. In our first approach, dealing with adaptive attackers is relatively easy as we are robustifying at operation time: if a change in the adversary's behavior is detected, it can be accounted for in the models used for the attacker's random utilities and probabilities, without retraining the algorithms. Indeed, if data about the adversary is available, models for random utilities and probabilities could be updated online in a Bayesian manner. 
In our second approach, we could retrain once we detect changes in attack patterns. However, note that in this approach learning (and robustification) is made over data minibatches. 
That is,
the attacker perturbs the first data minibatch with respect to the original defender model. Then, the defender retrains using this attacked minibatch, leading to a slightly more robustified model. Next, the attacker perturbs the second minibatch of data, with respect to the updated model (not the original one), so this attacker is also slightly more powerful, and so on.

}

\section{Conclusions}

\textcolor{black}{ Adversarial classification is an increasingly important problem 
within the emerging field of AML with relevance in numerous
 security, cybersecurity, law enforcement, and competitive business applications. 
The pioneering work by \cite{adversarialClassification2004} has framed, perhaps implicitly, most approaches in AC within a standard game-theoretic context, in spite of the unrealistic common knowledge assumptions required  (even questioned by those pioneers).}
On the other hand, in line with 
developments in robust Bayesian analysis \citep{ruggeri}, 
there have been several attempts in the Bayesian community to develop robust models, such as \cite{doi:10.1080/01621459.2018.1469995}. 
However, none of these approaches model explicitly the presence of adversaries and consequently would not perform properly in adversarial setups.

We have proposed a general Bayesian framework for adversarial classification that models explicitly the presence of an adversary and our uncertainty about his decision-making process. It is general in the sense that application-specific assumptions are kept to a minimum. A key ingredient required by our framework is the ability to sample from the distribution of originating instances given the (possibly attacked) observed one. 
For this, we first introduced AB-ACRA, a sampling scheme that leverages ARA and ABC to explicitly model the adversary's knowledge and interests, adding our uncertainty about them, and mitigating strong common knowledge assumptions prevalent in the literature. In large-scale problems, this approach easily becomes computationally expensive and we have presented an alternative 
proposal for differentiable, probabilistic classifiers.
In it, the computational load is moved 
to the training phase, simulating attacks from an adversary using the ARA approach,
and then adapting the training framework to obtain a classifier robustified against such attacks.

The proposed methods have performed effectively in the experiments considered. Bayesian methods seem indeed of high relevance to the AML community, since the uncertainties predicted by the models can be used to assess if an instance has been attacked (as a drift in the data distribution would suggest).
As an example, \cite{lakshminarayanan2017simple} proposed a baseline to get predictive uncertainties in large neural models using deep ensembling and AT to smooth those predictive estimates. However, they do not evaluate their framework against adversarial attacks. Our scalable framework showcases
the advantages of doing so. First, by estimating uncertainties  (e.g. using SG-MCMC algorithms) much better 
principled than their deep ensembles counterpart, as they target the usual posterior distribution in the Bayesian paradigm. Second, by generalizing AT to further improve robustness, we duly take into account the uncertainties faced by the attacker, inspired by the ARA framework. Besides, our experiments have suggested a
relevant regularizing effect.

Numerous lines for further research are worth pursuing
in this arena. We highlight four of relevance 
to the statistical community. First, we have just considered integrity violation attacks. Extensions to availability violation attacks, whose goal is to increase the wrong classification rate, would be important. 

\textcolor{black}{
Second, the  AB-ACRA scheme could be improved in several ways. We 
have introduced a vanilla ABC version exclusively; integrating recent advancements in ABC methods into the suggested probabilistic framework for AML presents a compelling avenue for further research which could alleviate the AB-ACRA computational bottleneck.
For instance, better sampling strategies proposed in the ABC literature could be adapted to our specific context, such as \cite{bortot2007inference}. Moreover, exploring how to build relevant summary statistics for our algorithm, e.g.\ by following ideas in \cite{fearnhead2012constructing}, could be fruitful. 
Finally, it is important to note our focus on Monte Carlo ABC. As demonstrated by \cite{papamakarios2016fast}, there are situations where learning an accurate parametric representation of the entire true posterior distribution requires fewer model simulations than what Monte Carlo ABC methods demand for producing a single sample from an approximate posterior. Therefore, exploring the adaptation of parametric approaches for likelihood-free inference to our AML framework could be of interest.}

Third, the approach for differentiable  models could be improved as well.
   Since it requires an SG-MCMC method to simulate attacks, instead of the vanilla SGLD sampler, we could use more efficient samplers, such as those introduced in \cite{gallego2018sgmcmc}. 
Finally, we have only touched upon classification 
problems but the ideas may be used in other areas
like standard regression or autoregressive tasks such as time series analysis or natural language processing.
\textcolor{black}{ Other relevant areas include considering non maximum expected utility 
adversaries, as with the prospect theory models presented in  decision analytic contexts in \cite{jesus}, and integrating our proposals with standard robust 
Bayesian analysis tools mentioned above}.

\textcolor{black}{
From a computational perspective, the previous sections illustrate the efficacy of the ARA approach in protecting  
classification algorithms. However, 
the aforementioned framework essentially simulates the attacker problem to forecast attacks and utilizes this information to optimize the defender's decision. This entails a non-trivial amount of computational resources and effort, in limited supply 
in many real-time environments. 
Therefore, future ARA research of an algorithmic nature is crucial.  An 
approach based on augmented probability simulation  \citep{EKIN2022} is promising in that it combines Monte Carlo sampling and optimization routines. However, should this approach prove computationally infeasible as well, alternative means may be required. Such alternatives may include expedient heuristics for the attacker's problem or approximation techniques that regress the attacker's best response function in a metamodeling sense.
}

All in all, we would finally stress that AML 
research has remained largely unexplored by the statistical community, remaining mostly within
the computer science domain. We  hope
that this paper will stimulate research oriented
towards leveraging powerful statistical tools to
tackle a problem of major relevance in modern
societies.



 \subsection*{Acknowledgments}
Work supported by the Severo Ochoa Excellence Programme SEV-2015-0554, the European Union's Horizon 2020 Research and Innovation Program under Grant Agreements 815003 (Trustonomy) and
101021797 \\
(STARLIGHT), the FBBVA project AMALFI  as well as NSF grant DMS-1638521 at SAMSI.
We acknowledge Virus Total for providing the malware dataset.
DRI is supported by the AXA-ICMAT Chair and the Spanish Ministry of Science program 
PID2021-124662OB-I00.  AR was supported by project RTC-2017-6593-7.  RN acknowledges support from grant FPU15-03636. VG acknowledges support from grant FPU16-05034 and PTQ2021-011758.
Part of the work was performed during the visit of VG, RN, DRI and FR to SAMSI (Statistical and Applied Mathematical Sciences Institute), Durham, NC, USA, within the "Games and Decisions in Risk and Reliability" program.

\bibliographystyle{imsart-nameyear}
\bibliography{references}


\end{document}